\PassOptionsToPackage{table}{xcolor}
\documentclass{article}

\usepackage{microtype}
\usepackage{graphicx}
\usepackage{booktabs} 

\usepackage{hyperref}
\usepackage{url}
\usepackage{subcaption}
\usepackage{xcolor}
\usepackage{adjustbox}
\usepackage{multirow}
\usepackage{tikz}
\usetikzlibrary{positioning, shapes, arrows.meta, calc, fit, backgrounds, shadows}

\definecolor{softgray}{gray}{0.55}
\definecolor{zerogray}{gray}{0.75}
\newcommand{\heatcell}[2]{%
  \pgfmathtruncatemacro{\isAbove}{#1 >= #2 ? 1 : 0}%
  \ifnum\isAbove=1
    \pgfmathsetmacro{\hintensity}{min((#1-#2)/#2*40, 40)}%
    \edef\hcspec{orange!\hintensity}%
  \else
    \pgfmathsetmacro{\hintensity}{min((#2-#1)/#2*40, 40)}%
    \edef\hcspec{teal!\hintensity}%
  \fi
  \expandafter\cellcolor\expandafter{\hcspec}#1%
}
\newcommand{\heatcellbf}[2]{%
  \pgfmathtruncatemacro{\isAbove}{#1 >= #2 ? 1 : 0}%
  \ifnum\isAbove=1
    \pgfmathsetmacro{\hintensity}{min((#1-#2)/#2*40, 40)}%
    \edef\hcspec{orange!\hintensity}%
  \else
    \pgfmathsetmacro{\hintensity}{min((#2-#1)/#2*40, 40)}%
    \edef\hcspec{teal!\hintensity}%
  \fi
  \expandafter\cellcolor\expandafter{\hcspec}\textbf{#1}%
}
\newcommand{\deltacol}[1]{%
  \pgfmathtruncatemacro{\dval}{#1}%
  \pgfmathtruncatemacro{\disPos}{\dval > 0 ? 1 : 0}%
  \ifnum\disPos=1
    \pgfmathsetmacro{\dintensity}{min(\dval * 6, 100)}%
    \edef\dcspec{orange!\dintensity!softgray}%
    {\expandafter\color\expandafter{\dcspec}+\dval}%
  \else
    \pgfmathtruncatemacro{\disNeg}{\dval < 0 ? 1 : 0}%
    \ifnum\disNeg=1
      \pgfmathtruncatemacro{\dabsval}{-\dval}%
      \pgfmathsetmacro{\dintensity}{min(\dabsval * 6, 100)}%
      \edef\dcspec{teal!\dintensity!softgray}%
      {\expandafter\color\expandafter{\dcspec}$-$\dabsval}%
    \else
      {\color{zerogray}0}%
    \fi
  \fi
}
\newcommand{\deltacolbf}[1]{%
  \pgfmathtruncatemacro{\dval}{#1}%
  \pgfmathtruncatemacro{\disPos}{\dval > 0 ? 1 : 0}%
  \ifnum\disPos=1
    \pgfmathsetmacro{\dintensity}{min(\dval * 6, 100)}%
    \edef\dcspec{orange!\dintensity!softgray}%
    {\expandafter\color\expandafter{\dcspec}\textbf{+\dval}}%
  \else
    \pgfmathtruncatemacro{\disNeg}{\dval < 0 ? 1 : 0}%
    \ifnum\disNeg=1
      \pgfmathtruncatemacro{\dabsval}{-\dval}%
      \pgfmathsetmacro{\dintensity}{min(\dabsval * 6, 100)}%
      \edef\dcspec{teal!\dintensity!softgray}%
      {\expandafter\color\expandafter{\dcspec}\textbf{$-$\dabsval}}%
    \else
      {\color{zerogray}\textbf{0}}%
    \fi
  \fi
}

\usepackage[preprint]{icml2026}
\makeatletter
\renewcommand{\ICML@preprint}{\textit{Preprint.}}
\makeatother

\usepackage{amsmath}
\usepackage{amssymb}
\usepackage{mathtools}
\usepackage{amsthm}
\usepackage{pifont}

\usepackage[capitalize,noabbrev]{cleveref}

\theoremstyle{plain}

\theoremstyle{definition}

\theoremstyle{remark}

\usepackage[textsize=tiny]{todonotes}

\icmltitlerunning{Input-Time Scaling: Adding Noise and Irrelevance for Reasoning}

\begin{document}

\twocolumn[
  \icmltitle{Input-Time Scaling: Adding Noise and Irrelevance into Less-Is-More Drastically Improves Reasoning Performance and Efficiency}



  \icmlsetsymbol{equal}{*}
  \icmlsetsymbol{dagger}{\textdagger}
  \icmlsetsymbol{co}{\S}

  \begin{icmlauthorlist}
    \icmlauthor{Rapheal (Yuming) Huang}{ir,co,dagger}
    \icmlauthor{Weilong Guo}{ir,co}
  \end{icmlauthorlist}
  \icmlaffiliation{ir}{Independent Researcher}

  \icmlcorrespondingauthor{Rapheal (Yuming) Huang}{raphealhuang@163.com}

  \icmlkeywords{Machine Learning, Reasoning, Input-Time Scaling, Less-Is-More}

  \vskip 0.3in
]



\printAffiliationsAndNotice{%
  \textbf{Co-authors (\S)} and \textbf{Corresponding Author (\textsuperscript{\textdagger}).} \textbf{Rapheal (Yuming) Huang} (\texttt{raphealhuang@163.com}) designed and implemented the pipelines, trained and evaluated models, gained and analyzed the results. He also contributed to manuscript writing, visualization, and intensive review \& refining. \textbf{Weilong Guo} (\texttt{guoweilong19@mails.ucas.ac.cn}) mainly contributed to aspects of experiments, data analysis, manuscript review and refining processes
}

\begin{abstract}
Large Language Models (LLMs) excel at reasoning, traditionally requiring high-quality large-scale data and extensive training. Recent works reveal a very appealing Less-Is-More phenomenon where very small, carefully curated high-quality datasets match resource-intensive approaches. In this work, we further systematically relax their quality constraints by adding controlled noise via persona context relevance and comparing datasets of different qualities. Counterintuitively, we find that mixing relevant and irrelevant contexts consistently across training and inference stages yields optimal results---a phenomenon we term training-testing co-design. Dataset quality comparisons show that high-quality data benefits weaker models on easy questions, while low-quality data achieves higher scores on hard questions with capable models. Across our experiments, reasoning performance is linked to reasoning efficiency. We, for the first time, found adding noisy and irrelevant contexts into queries can improve reasoning efficiency without any prices and targeted designs. Building on these insights, we propose Input-Time Scaling: applying small, low-quality data to capable models with training-testing co-design. This maintains Less-Is-More while further removing labor-intensive quality curation and improving reasoning effectiveness and efficiency, making the approach more applicable and affordable. Our method achieves 76.7\% pass@1 on AIME24/25 using Qwen2.5-32B-Instruct, and 90.0\%/80.0\% with DeepSeek-R1-Distill-Qwen-32B---state-of-the-art among Qwen2.5-32B variants. We are open-sourcing our datasets, pipelines, evaluation results, and checkpoints to facilitate reproducibility and further research.
\end{abstract}

\section{Introduction}

Mathematical reasoning in large language models has improved rapidly in the past few years \cite{ref14, ref42, ref18, ref51}, but the dominant recipe for getting there---extensive supervised fine-tuning on carefully filtered corpora, often followed by a reinforcement-learning stage---is expensive to run and labor-intensive to produce \cite{ref2, ref5, ref12, ref25}. The \textbf{Less-is-More} line of work \cite{ref12, ref37} has shown that a few thousand hand-picked examples can already match training runs orders of magnitude larger, yet the filtering effort underneath that success has not actually gone away; it has merely moved earlier in the pipeline. A natural next question is therefore how much of that filtering effort is genuinely load-bearing for reasoning, and what relaxing it would reveal about how models arrive at their reasoning abilities in the first place.

To answer this question, we need a way to vary data quality in a controlled manner rather than in the all-or-nothing way that filtering pipelines usually produce. We do so via \emph{context relevance}: content unrelated to a query degrades its quality in a semantically meaningful, adjustable way. Concretely, we adapt meta-cognition methods~\cite{ref47, ref36, ref70, ref69, ref28} that prompt an LLM to generate personas, but we use them differently from prior persona work~\citep{ref50, ref69}: instead of treating personas as scaffolding that is discarded before training, we generate personas at three levels of query-relevance (similar, random, dissimilar) and keep them in the input at both training and inference time, without modifying the query or its answer. The resulting datasets sit along a smooth noise axis, and by further comparing a fully curated base set with a minimally filtered one, we can also dissociate curation quality from curation quantity. If conventional quality heuristics hold, relevant personas and high-quality bases should consistently win; we find they do not.

From our sweep of these conditions across eight models spanning three families (Qwen2.5, DeepSeek-R1-Distill, Llama3) and five benchmarks, a coherent picture emerges: when the same kind of irrelevant context is present at both training and inference, models perform substantially better than when it appears at only one phase or at neither. We read this as evidence that context-induced shortcuts are what limits generalization under standard curation, and that denying those shortcuts at both phases is what unlocks the gains. Turning this observation into a practical recipe, we describe \textbf{Input-Time Scaling}: SFT on only 1K minimally curated examples, no RL, with persona context retained at both training and inference, applied to already-capable base models---state-of-the-art on AIME24/25 among 32B systems we compare against, with no reinforcement-learning stage.

\textbf{What this paper reports.}

\textit{Two results that directly test common assumptions in the literature:}
\begin{enumerate}
    \item[\textbf{(a)}] \textbf{Phase-consistent irrelevant context is the primary driver:} The largest improvements we see come from injecting persona context in \emph{both} training and inference, rather than in either phase alone; the specific persona type (relevant, random, or dissimilar) has only secondary effect. This suggests the benefit is not explained by viewing persona as ``useful extra data''---what matters is that the model sees the same kind of context at train and test time.
    \item[\textbf{(b)}] \textbf{Quality vs.\ quantity inverts with capacity:} Capable base models get their highest ceilings from the \emph{least} filtered data in our comparison, contrary to the common ``tighter curation is always better'' heuristic. Holding the distribution fixed and enlarging the training set from 1K to 15K examples actually \emph{reduces} performance by roughly 15\%, consistent with duplication reinforcing spurious patterns rather than building generalization.
\end{enumerate}
\textit{Three observations that surfaced from closer analysis of the same runs:}
\begin{enumerate}
    \item[\textbf{(c)}] \textbf{Efficiency co-emerges with accuracy, unoptimized:} Configurations that perform better also produce shorter reasoning traces, even though our training objective rewards neither brevity nor efficiency. This compression appears as a side effect rather than a direct optimization target.
    \item[\textbf{(d)}] \textbf{Reduced benchmark-specific overfitting:} Baseline methods typically exhibit a 10--20 point gap between AIME24 and AIME25 that is most parsimoniously explained by partial test-set leakage. Our configurations substantially shrink this gap while also raising absolute scores on both benchmarks, pointing to representations that are less tied to any specific benchmark distribution.
    \item[\textbf{(e)}] \textbf{Capacity-gated emergence:} The benefit of irrelevant context is essentially absent at small scales and appears sharply only at larger scale on harder problems. The effect is therefore a scaling phenomenon gated by capacity and difficulty, not a uniform training trick.
\end{enumerate}

\textbf{A single account for all five.} We read these findings through one lens: \emph{context that cannot be turned into a shortcut pushes the model toward reasoning policies that generalize}. Under this reading, (a) follows because noisy context at both training and inference blocks shortcut routes at both phases, leaving only transferable reasoning; (b) follows because over-curated data invites shortcut formation, and duplicating it amplifies the trap; (c) follows because shortcut-heavy solutions tend to be verbose and template-matched, so moving away from them naturally shortens traces; (d) follows because a model that has not anchored to benchmark-specific cues performs more uniformly across closely related benchmarks; and (e) follows because learning general policies from noisy inputs is harder than memorizing shortcuts from clean ones, so the effect only surfaces once the base model is capable enough to do so.

\textbf{The resulting recipe.} Turning these observations into a practical setup, we describe \textbf{Input-Time Scaling}: SFT on only 1K minimally curated examples, no RL, with persona context retained consistently at both training and inference, applied to already-capable base models. On AIME24/25 this yields 76.7\% pass@1 with Qwen2.5-32B-Instruct and 90.0\%/80.0\% with DeepSeek-R1-Distill-Qwen-32B---competitive with systems that use substantially larger bases or far more training data. Datasets, pipelines, evaluation scripts, and checkpoints will be released.

\begin{figure*}[!htbp]
    \centering
    \includegraphics[width=5.5in, height=2in]{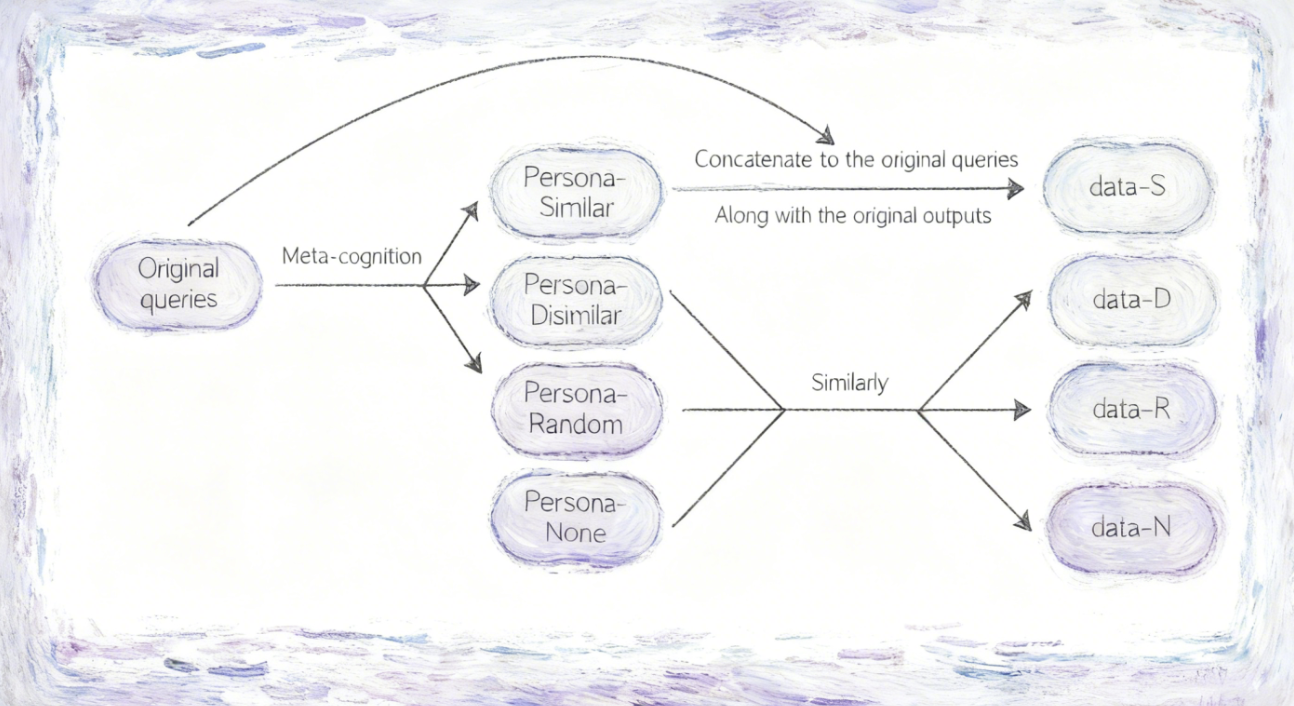}
    \caption{Data processing pipeline overview. We use meta-cognition methods to create three kinds of persona contexts concatenated to queries. For each persona type, we generate corresponding persona contexts from the original queries, then concatenate these contexts to the queries. The concatenated queries plus the original cot\&answers form the corresponding dataset. Note that each query receives a unique persona context with no reuse. Overall, we create four datasets: three with different persona types and one with the original unchanged dataset. By controlling the relevance between personas and queries, we control the noise levels added to the queries.}
    \label{fig:data_process}
\end{figure*}

\section{Methods: Systematically Probing Quality and Noise in Reasoning Data}
\label{methods}
As motivated in Section~1, we use persona-query relevance as a controllable noise axis and compare datasets of different quality levels. This yields two experimental dimensions: \textbf{(1) Noise level}, controlled by the type of persona concatenated to each query (Section~\ref{sec:persona-strategies}); and \textbf{(2) Data quality}, controlled by the choice of base dataset (Section~\ref{sec:dataset-quality}). Crossed with training and testing phases, these dimensions produce a rich experimental grid. We provide a reproducibility statement in Appendix~\ref{reproducibility}.

\subsection{Context Strategies: Controlling Noise via Relevance}
\label{sec:persona-strategies}

We use LLM-generated personas as our noise injection mechanism to control the contexts. Unlike statistical noise (e.g., embedding perturbation) or data augmentation methods that discard personas after generation, our personas are \emph{structured, semantically coherent text} that is retained throughout both training and inference. For each query, we prompt DeepSeek-R1-0528 to generate a persona context, then concatenate it to the original input while leaving the query-answer pair unmodified. The relevance between the persona and the query determines the noise level. This yields four strategies:

\begin{enumerate}
    \item[-] \textbf{Persona-None (N)}: Keep the original input without modifications (baseline).
    \item[-] \textbf{Persona-Similar (S)}: Generate a persona related to the query. This provides potentially helpful background context.
    \item[-] \textbf{Persona-Dissimilar (D)}: Generate a persona unrelated to the query that cannot provide any useful information. This introduces pure irrelevant noise.
    \item[-] \textbf{Persona-Random (R)}: Randomly choose a domain via meta-cognition and generate a corresponding persona. The domain may or may not relate to the query, producing a mixture of relevant and irrelevant noise.
\end{enumerate}

These four strategies are applied independently during training and testing, producing a 4$\times$4 grid of 16 train-test combinations per model-dataset pair. A naive reading of conventional quality heuristics would predict that S should help, D should hurt, and R should fall in between. If D or R match or exceed S, that prediction is falsified and the assumption that irrelevant context can only degrade reasoning no longer stands. See Appendix~\ref{prompt} for detailed prompts and Figure~\ref{fig:data_process} for a visualization of the data-processing pipeline.

\subsection{Dataset Quality: High vs.\ Low Curation Effort}
\label{sec:dataset-quality}
We compare two datasets representing opposite ends of the quality spectrum. \textbf{LIMO}~\cite{ref12} ($<$1K pairs) represents high quality: queries are carefully selected for difficulty and diversity, reasoning chains are curated, and answers are verified. \textbf{OpenThought (OT)}~\cite{ref14} ($>$1M pairs) represents low quality: there is no CoT or answer filtering---many entries lack answers entirely---and each query is sampled 16 times via QwQ-32B, resulting in lower query diversity than LIMO. We randomly sample 1K examples from OT to match LIMO's scale, isolating quality as the variable rather than dataset size. If quality heuristics hold, LIMO should consistently outperform OT across all model scales and persona strategies.

\section{Experiments}
\label{experiments}

\subsection{Models and Baselines}
We experiment across 3 model families and 8 model sizes to study how noise effects interact with model capacity: \textbf{Qwen2.5}~\cite{ref56} (3B, 7B, 32B) as our primary testbed of widely-used non-reasoning models, \textbf{DeepSeek-R1-Distill-Qwen}~\cite{ref42} (1.5B, 7B, 32B) to test on reasoning-specialized variants, and \textbf{Llama}~\cite{ref68} (3.1-8B, 3.2-3B) for cross-family validation. We primarily compare against LIMO~\cite{ref12}, S1~\cite{ref37}, OpenThinker~\cite{ref14}, MiroMind-M1~\cite{ref18}, and Skywork-OR1~\cite{ref51} at the 32B scale where our main claims are strongest.

\subsection{Training Setup}
\label{sec:training-setup}

For each base dataset (LIMO and OT), we create 4 training variants by applying the persona strategies from Section~\ref{sec:persona-strategies} (N, S, D, R) to all queries, while keeping the original CoT and answers unchanged. For OT, we sample 1K examples from the math category, filtering only for entries containing ``final answer'' and ``boxed\{'' in their outputs; no other quality filtering is applied. Each model is trained on all 8 variants (2 datasets $\times$ 4 strategies).

We train using 360-LlamaFactory~\cite{ref53} (a variant of LlamaFactory~\cite{ref54} with sequence parallelism). All models use a cutoff length of 20,000 tokens without packing, 240 update steps, batch size 48, learning rate 5e-6 with cosine schedule. We also explore model merging via mergekit~\cite{ref55} and majority voting across three models; see Appendix~\ref{merge_majority} for these results. We provide example training scripts in the supplementary materials, with full training details in Appendix~\ref{reproducibility}.

\subsection{Evaluation Setup}

\textbf{Benchmarks:} We use AIME24 and AIME25~\cite{ref58} as primary benchmarks for hard mathematical reasoning, supplemented by MATH500~\cite{ref71} and GPQA~\cite{ref64}. Each test set is also augmented into 4 variants (N, S, D, R), following the same persona process as training (Figure~\ref{fig1}).

\textbf{Protocol:} Each trained model (8 per base model) is evaluated on 4 test variants per benchmark, yielding 32 evaluations per base model per benchmark and 256+ evaluation sets in total. For computational efficiency, 32B models use greedy decoding (pass@1); smaller models use pass@1 averaged over 4 samples, following the LIMO protocol.

\textbf{Reading the results:} Throughout the paper, we use the notation \textbf{X-Y} to denote training strategy X and testing strategy Y (e.g., ``S-D'' means training with Similar personas, testing with Dissimilar). We report ``Avg'' (mean across all four benchmarks) and ``Avg2'' (mean of AIME24 and AIME25 only), as these hard reasoning benchmarks exhibit performance patterns distinct from the supplementary datasets. Table~\ref{tab:qwen2.5-32b} shows a complete example of this 4$\times$4 grid for both LIMO and OT datasets.

\begin{table*}[t]
\centering
\setlength{\tabcolsep}{0.7mm}{
\begin{tabular}{c|c|cccccc|cccccc}
\hline
 &  & \multicolumn{6}{c|}{\textbf{LIMO (High Quality)}} & \multicolumn{6}{c}{\textbf{OT (Low Quality)}} \\
\cline{3-14}
\textbf{Train} & \textbf{Test} & \textbf{A24} & \textbf{A25} & \textbf{MATH} & \textbf{GPQA} & \textbf{avg} & \textbf{avg2} & \textbf{A24} & \textbf{A25} & \textbf{MATH} & \textbf{GPQA} & \textbf{avg} & \textbf{avg2} \\
\hline
\multirow{4}{*}{\textbf{N}}
 & \textbf{N} & 61.00 & 40.00 & 92.00 & 53.54 & 61.64 & 50.50 & 63.67 & 50.00 & 94.80 & 66.67 & 68.79 & 56.84 \\
 & \textbf{R} & 43.33 & 40.00 & 90.20 & 51.01 & 56.14 & 41.67 & 60.00 & 53.33 & 91.20 & 66.67 & 67.80 & 56.67 \\
 & \textbf{S} & \textbf{63.33} & 40.00 & 89.00 & 50.00 & 60.58 & 51.67 & 60.00 & 56.67 & 90.00 & 64.85 & 67.88 & 58.34 \\
 & \textbf{D} & 40.00 & 36.67 & 87.20 & 51.52 & 53.85 & 38.34 & 60.00 & 60.00 & 89.40 & 61.62 & 67.76 & 60.00 \\
\hline
\multirow{4}{*}{\textbf{R}}
 & \textbf{N} & 59.33 & 36.67 & \textbf{93.80} & \textbf{68.69} & 64.62 & 48.00 & 40.00 & 26.67 & 92.80 & 68.18 & 56.91 & 33.34 \\
 & \textbf{R} & \textbf{63.33} & 50.00 & 93.20 & 65.66 & 68.05 & 56.67 & 60.00 & 70.00 & 94.80 & 69.70 & 73.63 & 65.00 \\
 & \textbf{S} & \textbf{63.33} & 53.33 & 93.40 & 66.34 & \textbf{69.10} & \textbf{58.33} & 73.33 & 53.33 & 94.00 & 67.33 & 72.00 & 63.33 \\
 & \textbf{D} & 53.33 & \textbf{56.67} & 92.40 & 63.13 & 66.38 & 55.00 & 66.67 & 56.67 & 94.40 & 69.70 & 71.86 & 61.67 \\
\hline
\multirow{4}{*}{\textbf{S}}
 & \textbf{N} & 60.33 & 46.67 & 92.20 & 59.09 & 64.57 & 53.50 & 43.33 & 33.33 & 91.60 & 68.69 & 59.24 & 38.33 \\
 & \textbf{R} & 43.33 & 43.33 & 90.20 & 58.08 & 58.74 & 43.33 & 73.33 & 60.00 & 94.40 & 71.72 & 74.86 & 66.67 \\
 & \textbf{S} & 60.00 & 50.00 & 92.00 & 56.44 & 64.61 & 55.00 & 66.67 & 66.67 & 94.60 & 69.31 & 74.31 & 66.67 \\
 & \textbf{D} & 53.33 & 50.00 & 89.40 & 59.09 & 62.96 & 51.67 & \textbf{76.67} & 70.00 & \textbf{95.00} & 66.67 & \textbf{77.09} & \textbf{73.34} \\
\hline
\multirow{4}{*}{\textbf{D}}
 & \textbf{N} & 55.67 & 43.33 & 92.20 & 58.59 & 62.45 & 49.50 & 33.33 & 36.67 & 91.80 & 66.16 & 56.99 & 35.00 \\
 & \textbf{R} & 60.00 & 50.00 & 89.00 & 53.03 & 63.01 & 55.00 & 63.33 & 66.67 & 94.60 & \textbf{72.22} & 74.21 & 65.00 \\
 & \textbf{S} & \textbf{63.33} & 46.67 & 90.60 & 57.43 & 64.51 & 55.00 & 70.00 & \textbf{73.33} & 94.20 & 67.82 & 76.34 & 71.67 \\
 & \textbf{D} & 53.33 & 33.33 & 88.00 & 52.02 & 56.67 & 43.33 & 66.67 & 60.00 & 94.20 & 67.68 & 72.14 & 63.34 \\
\hline
\end{tabular}}
\caption{\textbf{qwen2.5-32B-Instruct results:} "A24" is short for AIME24, and "A25" is short for AIME25. "Avg" is the average of four dataset results, and "Avg2" only counts AIME24 and AIME25 results. We get the overall best performance with S-D under OT datasets, which is on average 8\% higher than the best performance (R-S) under LIMO datasets. The training-testing co-design is more obvious on OT datasets and performs better.}
\label{tab:qwen2.5-32b}
\end{table*}


\begin{figure*}[!htbp]
    \centering
    \begin{subfigure}[b]{0.48\textwidth}
        \centering
        \includegraphics[width=\textwidth]{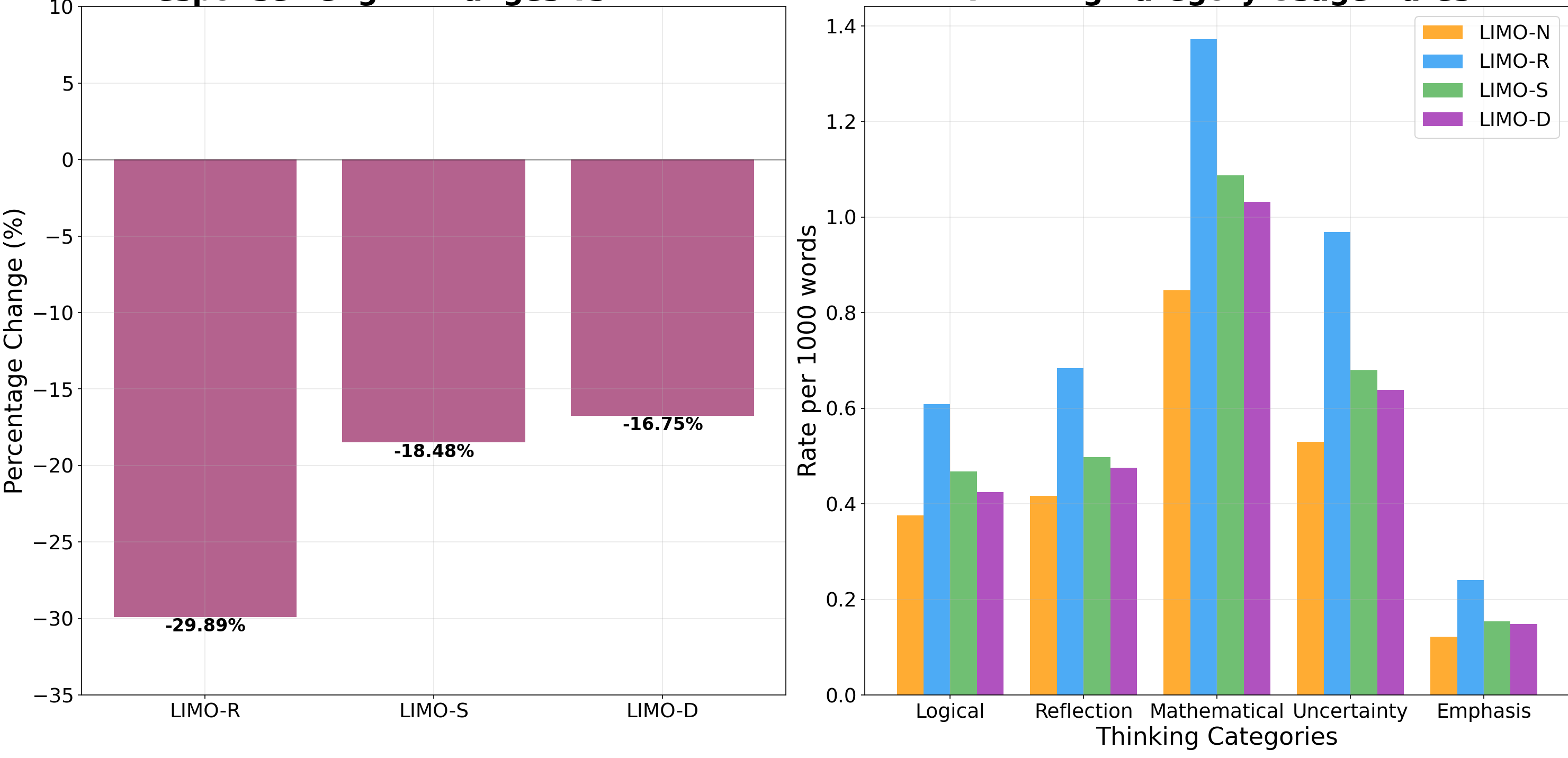}
        \caption{training, LIMO datasets.}
    \end{subfigure}
    \hfill
    \begin{subfigure}[b]{0.48\textwidth}
        \centering
        \includegraphics[width=\textwidth]{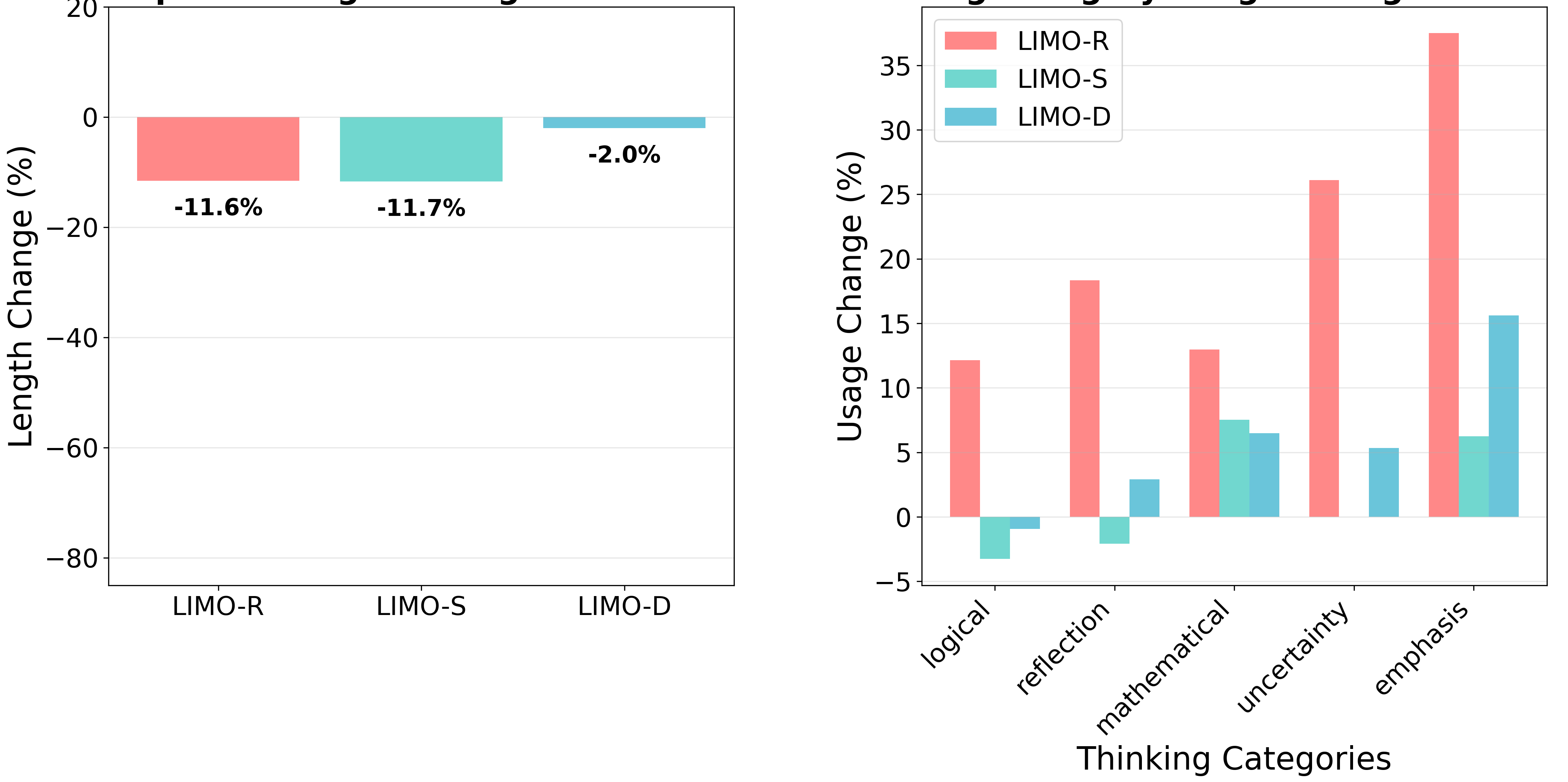}
        \caption{testing, LIMO datasets.}
    \end{subfigure}
    
    \vspace{0.3cm}
    
    \begin{subfigure}[b]{0.48\textwidth}
        \centering
        \includegraphics[width=\textwidth]{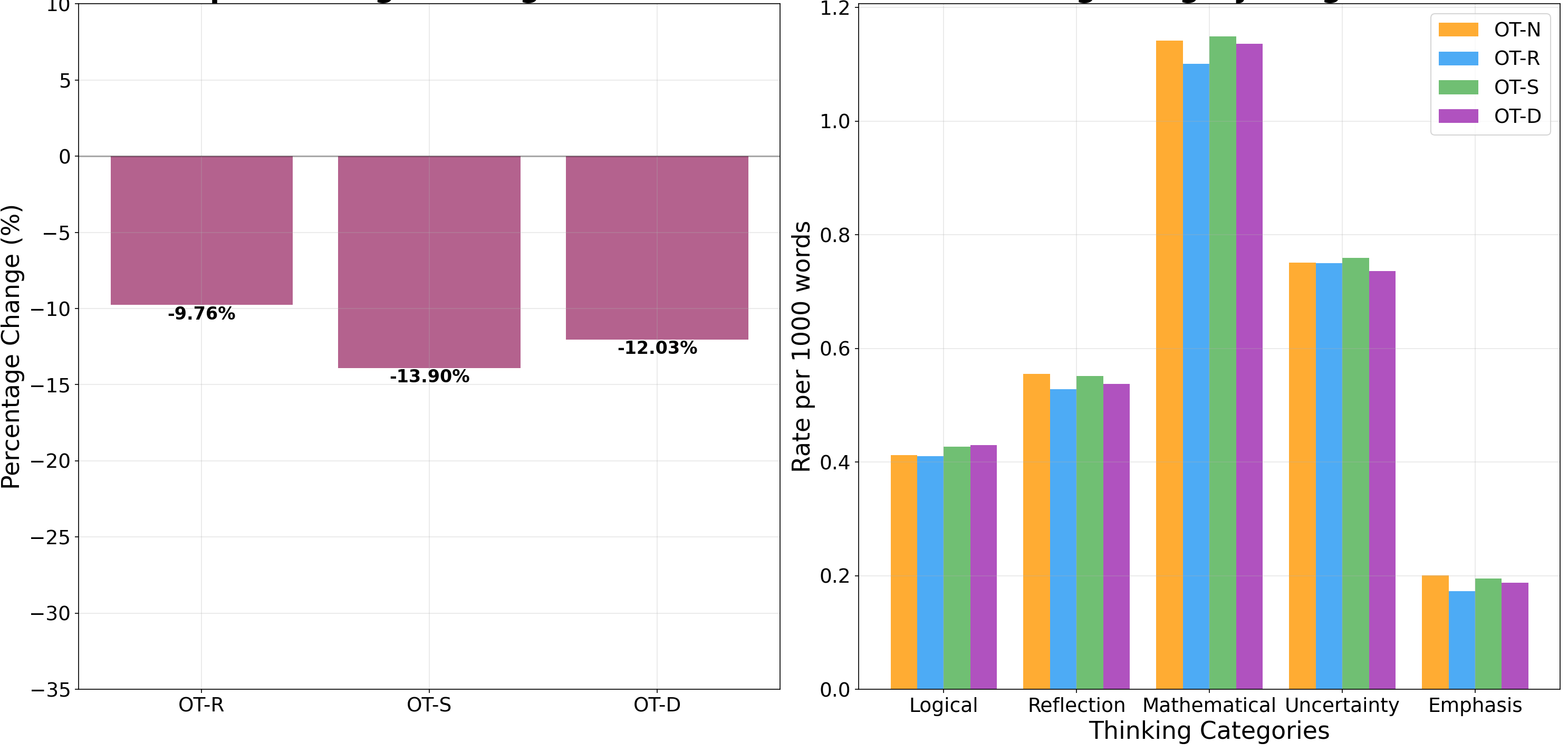}
        \caption{training, OT datasets.}
    \end{subfigure}
    \hfill
    \begin{subfigure}[b]{0.48\textwidth}
        \centering
        \includegraphics[width=\textwidth]{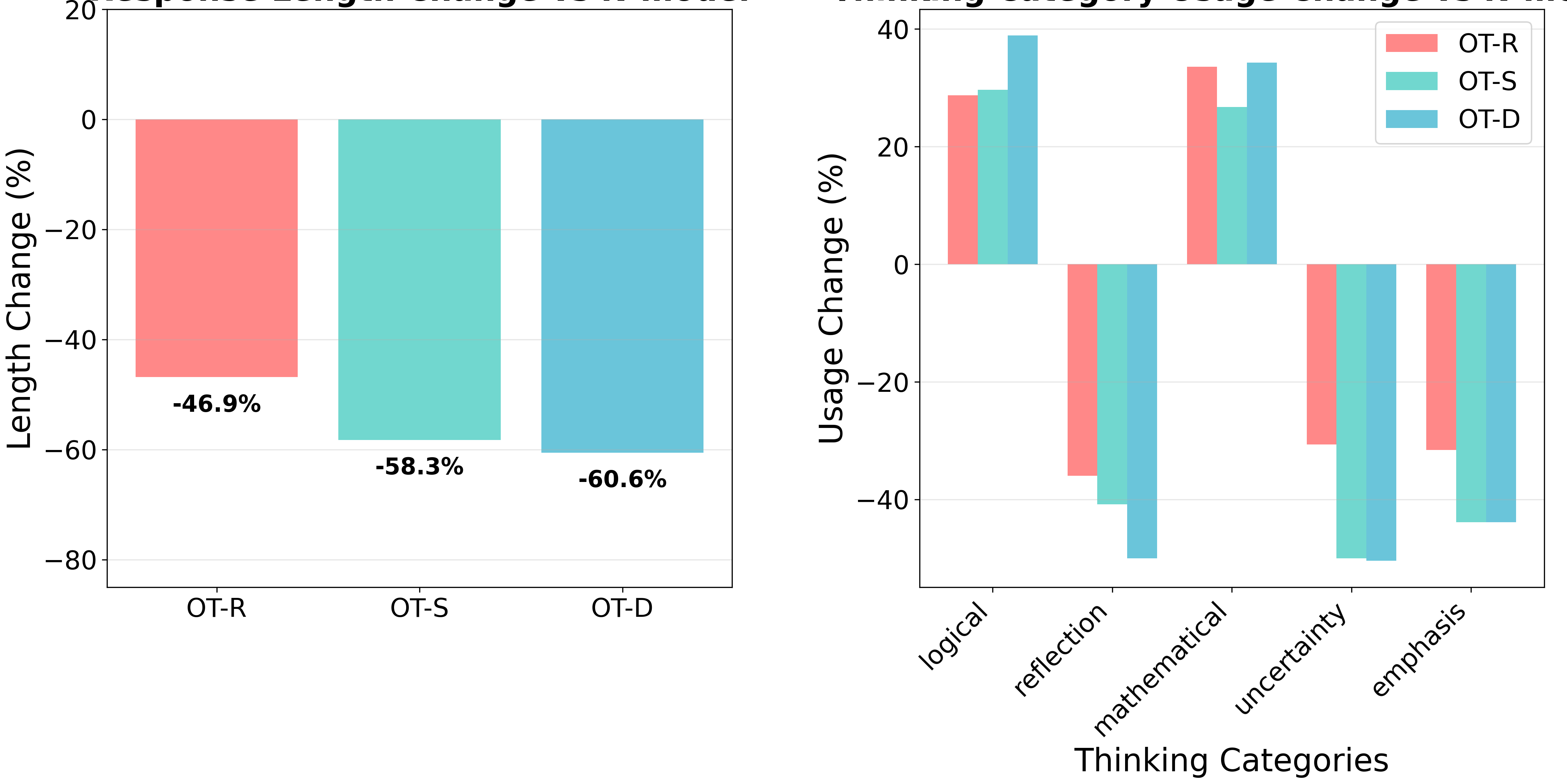}
        \caption{testing, OT datasets.}
    \end{subfigure}
    
    \caption{\textbf{Performance Comparison Between Training and Testing Strategies of Qwen2.5-32B-Instruct:} (a,c) are the training visualizations ; (b,d) are the testing visualizations. Each sub figure contains two part and both are compared to -N strategy. Left contains the output length change rates, the right is the visualizations into thinking categories. Applying any strategy during training or testing will shorten the output length, making the reasoning more efficient, which is linked to improved performance.}
    \label{fig:qwen-32b}
\end{figure*}


\section{Empirical Findings}

We report five empirical findings from our experiment sweep across eight models, three families, and five benchmarks. The first two results test common literature hypotheses directly: how noise affects reasoning, and whether conventional quality heuristics hold. The remaining three surfaced only under closer analysis: reasoning efficiency, benchmark-gap patterns, and a capacity-dependent emergence of the noise benefit. Section~\ref{sec:unified-summary} consolidates the picture across model families, and Section~\ref{sec:discussion} offers a unified interpretation.

\subsection{Testing the Noise Hypothesis}

\subsubsection{Train-Test Co-Design}
\label{sec:finding-a}

We examine the 4$\times$4 strategy grid (Table~\ref{tab:qwen2.5-32b}) to see how persona augmentation at training and testing shapes reasoning performance. Three observations emerge.

\emph{Phase matching matters more than persona choice.} On OT, configurations where persona is present in \emph{both} training and inference ([R,S,D]-[R,S,D]) beat configurations where it is present in only one phase by a large margin---roughly a mid-teens absolute-point gap on the four-benchmark average (and a larger gap on the AIME-only average) against the train-only setting, with smaller but still consistent gaps against the test-only and no-persona settings (Table~\ref{tab:qwen2.5-32b}, right). LIMO shows the same ordering at smaller magnitudes. We call this the \textbf{train--test co-design} effect.

\emph{Which persona is used is secondary.} Conditional on the phases matching, R, S, and D all reach similar AIME24/25 scores---the binary presence-in-both is what carries the effect, not the fine-grained relevance setting.

\emph{Noisy personas beat the relevant one at training time.} On LIMO, using noisy personas (R, D) at training beats both the no-persona baseline and the relevant S on the four-benchmark average. The reversal is not large in absolute terms, but it goes the wrong way for any view in which relevance is what helps.

These three observations hold across all eight models (Appendix~\ref{full-results}): R or D yield the best training configuration in a clear majority of runs, substantially more than the relevant S or the baseline N; S is the most reliable choice at test time, though R and D often produce higher absolute scores; and the matched-configuration advantage grows with model size. All persona types---including the explicitly irrelevant one---produce gains, consistent with the reasoning-policy mechanism we return to in Section~\ref{sec:discussion}.

\subsection{Testing the Quality Hypothesis}

\subsubsection{Relaxed Quality vs.\ Quantity}
\label{sec:finding-b}

\textbf{Low quality outperforms high quality at scale.} Comparing LIMO (high quality) and OT (low quality) on Qwen2.5-32B-Instruct (Table~\ref{tab:qwen2.5-32b}), OT achieves approximately +8\% avg and +15\% avg2 over LIMO. The two datasets also respond differently to noise: on LIMO, noise (R, D) helps only when combined with co-design, with best performance at 63.3\%/53.3\% on AIME24/25 (R-S). On OT, co-design with any persona strategy yields dramatically higher scores, with the best at 76.7\%/70.0\% (S-D)---on average 35\% higher than testing without personas. DeepSeek-R1-Distill-Qwen-32B (Table~\ref{tab:deepseek-r1-distill-qwen-32b}) shows a similar OT advantage ($\sim$10\% higher), though LIMO achieves its best scores on N-N for this model---an atypical pattern that reinforces the capacity-dependent nature of these effects.

\textbf{Scaling can hurt.} Does more data amplify the benefits? We compare OT-1K against OT-15K (15$\times$ more data from the same distribution) on Qwen2.5-32B-Instruct (Table~\ref{tab:ot_combined} in Appendix). OT-15K performs \textbf{15\% worse on average} across all strategy combinations: its best configuration (D-D: 66.7\%/60.0\%) falls far short of OT-1K's best (S-D: 76.7\%/70.0\%), and it shows no clear co-design patterns.

\subsection{Emergent Discoveries from Deeper Analysis}

The noise and quality findings above point to patterns beyond what we originally set out to test. Closer examination surfaces two additional phenomena: reasoning efficiency emerges as a free side effect, and the quality--capacity relationship is governed by a capacity-dependent threshold.

\subsubsection{Intrinsic Efficiency: Shorter Traces as a Side Effect}
\label{sec:finding-c}

To understand the mechanistic basis of co-design, we analyze token distribution patterns in model outputs, categorizing thinking tokens into five types: logic, reflection, mathematics, uncertainty, and emphasis (Appendix~\ref{token-categories}).

\textbf{During training}, persona strategies (R, S, D) consistently yield shorter responses than baseline (N) while maintaining or increasing thinking token proportions across all categories (Figure~\ref{fig:qwen-32b}a,c). Models express equivalent or richer reasoning in fewer tokens---improved efficiency that correlates with better performance. This generalizes across Qwen2.5-7B and DeepSeek-Distill-Qwen-7B on OT (Figure~\ref{fig:combined_comparison}), and DeepSeek-Distill-Qwen-32B on OT (Figure~\ref{fig:deepseek-32b}).

\textbf{During testing}, the same compression pattern emerges independently: persona strategies produce shorter, higher-performing responses (Figures~\ref{fig:qwen-32b}b,d and~\ref{fig:deepseek-32b}d).

\textbf{Notable exception:} DeepSeek-Distill-Qwen-32B and Qwen2.5-7B on LIMO show increased length with persona strategies, with optimal performance from Persona-N. This mirrors a model-specific incompatibility with the LIMO dataset rather than contradicting the efficiency principle: in every case, the best-performing configuration for a given model also produces the most efficient reasoning traces.

\subsubsection{Capacity-Gated Effect and Quality--Capacity Tradeoff}
\label{sec:finding-e}

The OT advantage is not universal---it depends on model capacity (Appendix~\ref{full-results}). For smaller models (1.5B--8B), high-quality LIMO matches or exceeds OT, because aggregate scores are dominated by easier benchmarks (e.g., MATH) where curated data provides a clear advantage. As model size increases to 32B, the performance gap on easy problems narrows, and the aggregate metric becomes dominated by hard benchmarks (AIME24/25) where OT excels. This reveals a fundamental \textbf{quality-capacity tradeoff}: lower-quality data extends the performance ceiling on hard tasks, but only when paired with sufficiently capable models.

\subsection{Unified Summary of Findings Across Model Families}
\label{sec:unified-summary}

Across all eight models, two patterns are clear. First, the co-design advantage is not uniform at small scales: at 1.5--8B the improvement over the no-persona baseline is mixed (roughly $-1$ to $+6$ points on the AIME average). Second, the advantage grows sharply with capacity, reaching double-digit point gains at 32B on both Qwen2.5 and DeepSeek. The same pattern is visible in the quality-vs-capacity comparison: at small scales the curated dataset often matches or exceeds the minimally filtered one, but at 32B the minimally filtered data clearly wins on hard benchmarks.

\section{Input-Time Scaling and Its Effectiveness}
\label{sec:its}
\textbf{Input-Time Scaling} packages the findings above into a practical recipe. Rather than treating data curation and inference-time prompting as independent problems, the method couples input augmentation across both training and testing phases: it is the coupling itself, not either phase alone, that produces the gains. Concretely, we fine-tune sufficiently capable base models on very small, minimally filtered datasets with persona contexts applied during both training and testing. In our setup this yields up to roughly 27 absolute points over no-persona baselines on hard AIME problems, without paying for expensive curation. Table~\ref{main-comparison} places the resulting models side-by-side with existing systems.

Four patterns in Table~\ref{main-comparison} are worth highlighting. \textbf{(1) Quality curation limits the ceiling:} Against same-budget methods (LIMO, S1; 1K SFT each) on Qwen2.5-32B-Instruct, relaxing the data-quality bar while injecting noise yields roughly +20 absolute points on AIME24/25. \textbf{(2) Scaling quantity hurts:} Against OpenThinker2-32B (1M examples, same base and SFT process), our method uses about $1000\times$ less data yet gains around +22 points on AIME25, consistent with naively scaling from the same distribution hurting rather than helping. \textbf{(3) Co-design replaces RL:} Against RL-based methods (Skywork-OR1, MiroMind; 62--124K RL data) on DeepSeek-R1-Distill-Qwen-32B, a single co-design configuration (D-S) reaches 86.7\%/80.0\% with only 1K SFT examples and no RL---roughly $+9$/$+12$ points over the RL baselines. \textbf{(4) Paradigm outweighs scale:} The resulting 32B model reaches or surpasses several much larger or frontier systems---DeepSeek-R1~\cite{ref42}, OpenAI-o1~\cite{ref59}, Grok-3-Beta~\cite{ref9}, Qwen3-235B-A22B~\cite{ref62}---on AIME24, suggesting that the train--test coupling is doing real work that additional scale alone does not provide.

\section{Discussion: A Unified Hypothesis for Noise-Driven Reasoning}
\label{sec:discussion}

Taken together, the five findings become coherent under one reading: \emph{training with noisy, irrelevant context pushes models toward reasoning policies that generalize, by making context-tied shortcuts unusable at both training and inference time.} We walk through each finding under this reading.

Consistent persona augmentation across training and inference is the single dominant factor for performance---far more than the specific persona type. Irrelevant augmentation matches or exceeds relevant augmentation: R and D yield optimal training performance in a majority of our runs, substantially more often than the semantically relevant S, and well above the no-persona baseline. Human intuition would predict that irrelevant personas should hurt, yet all types produce large improvements, suggesting that models do not simply extract useful background but instead treat any rich context as a cue to engage more carefully. Under our reading, this is expected: once the model has been trained against varied irrelevant context, seeing similar context at test time places it inside the distribution where its generalized policies operate, rather than triggering a distribution shift.

Conventional quality heuristics invert at scale in our setup: minimally filtered data reaches higher performance ceilings than the curated alternative on capable models, and enlarging the training set from 1K to roughly 15K under the same distribution actively reduces performance. This is consistent with heavily filtered data making shortcut formation easy---the model anchors on surface features of a homogeneous corpus---while noisier data resists such anchoring, and duplicating the same distribution only amplifies the anchor.

Higher performance is also consistently linked to shorter reasoning traces. Unlike prior work that explicitly optimizes for brevity~\cite{ref42, ref44}, this compression arises as a side effect of persona augmentation. The simplicity-based reading is natural: reasoning policies that generalize tend to be more compact than a library of problem-specific shortcuts, because a single general rule can cover many cases that a shortcut library must enumerate separately. A model that has extracted the underlying principle behind a class of problems can therefore answer them more directly.

On benchmark-gap patterns, the baseline methods we compared show AIME24-vs-AIME25 gaps roughly in the high-single-digit to high-teens range---a pattern that holds across methods we did not design, making a genuine 1-year difficulty difference hard to sustain as the explanation; the simpler reading is that earlier problems are more represented in pretraining corpora. Our configurations substantially shrink this gap while also raising absolute scores on both benchmarks, consistent with the model not having anchored on benchmark-specific surface cues.

On scaling, the advantage is not uniform across sizes. At 1.5--8B the improvement over the no-persona baseline is small and mixed on the AIME average (roughly a few points in either direction), while at 32B it reaches double-digit point gains on both Qwen2.5 and DeepSeek. Under our reading, extracting general policies from noisy inputs is strictly harder than memorizing shortcuts from clean ones, so the benefit should only surface once the base model has enough capacity to do it---which is the pattern we observe.

\textbf{Where the evidence is strongest.} Two of the observations carry more weight than the others. The reduced benchmark-gap pattern is hard to produce accidentally: it requires consistent behavior across closely related benchmark pairs, which is what a model with less benchmark-specific behavior should show. The capacity-dependent emergence pattern is similarly hard to explain as a statistical artifact, because it shows a sharp threshold in size rather than a smooth trend.

\textbf{Open questions.} Several directions remain: what learning dynamics explain why small input-side modifications produce large accuracy shifts; whether the same principle extends beyond mathematical reasoning; what the precise capacity threshold looks like; and how noise-augmented SFT interacts with reinforcement learning, including whether it can provide a better cold-start point for RL-based training.

\section{Conclusion}
Across our experiment sweep---eight models from three families, five benchmarks, and a 4$\times$4 train--test context grid per model-dataset pair---we observe that adding noisy, query-irrelevant context consistently improves LLM reasoning when the same kind of context is applied at both training and inference. The single-condition gains from either phase alone are small; the cross-phase coupling is where the improvement lives. Turning this into a practical recipe, we describe Input-Time Scaling: SFT on only 1K minimally curated examples, no RL, with persona context retained at both training and inference, applied to already-capable base models. On AIME24/25, this gives 76.7\% pass@1 on Qwen2.5-32B-Instruct and 90.0\%/80.0\% on DeepSeek-R1-Distill-Qwen-32B. The method is simple and the effect is visible across base families, suggesting the mechanism is general rather than model-specific. Open questions for future work include what learning dynamics drive the coupling, whether it transfers beyond mathematical reasoning, where the capacity threshold actually sits, and how the approach interacts with reinforcement-learning-based post-training.

\nocite{langley00}

\bibliography{example_paper}
\bibliographystyle{icml2026}

\newpage
\appendix
\onecolumn
\section{Appendix}
\subsection{Reproducibility Statement}
\label{reproducibility}
You can find the train\&test sets curation details in \ref{methods} part. The specific prompts can be found in the APPENDIX \ref{prompt}. Our complete pipeline is to curate the corresponding train\&test sets, using SFT as shown in the \ref{experiments} part, and evaluate using the curated test sets. There is no filtering during the whole process, and all samples are selected randomly as mentioned. 

We provide example files (in our \textbf{supplementary zip file}) contains the process we use to create personas, and some dataset examples. \textbf{data\_process\_final.ipynb}: contains full process of creating the personas. We contain some example outputs of the personas. \textbf{json data files}: contains example training data for limo datasets with three (S, D, R) persona strategies. \textbf{sft\_LIMO.sh}: contains the example training script we use to train our models with 360-LlamaFactory. The efficient batch size depends on the accumulations and parallel settings. Overall we have a batch size about 48 and training steps about 240. \textbf{BoN\_final.py}: Provides a toy example of our naive BoN test. We only consider more than 2 of the 3 answers are consistant and correct, then we count the BoN result of the answers is correct.

\subsection{Related Works}
\label{sec:related}

Current large language models \cite{ref14, ref18, ref42, ref51} have achieved impressive performance on math and other reasoning tasks. They are usually post-trained on carefully curated large-scale datasets (data and training scaling), following a two-stage training pipeline that relies on different intuitive heuristics (inductive biases) to guide filtering \cite{ref2, ref5, ref12, ref25}. Quality and diversity are both considered important for LLMs, but defining quality is non-trivial; diversity is relatively more objective \cite{ref2, ref5, ref14}. Some methods \cite{ref12, ref14, ref38, ref37, ref40} treat difficulty as a more operational selection criterion: queries that contain more steps, require more information, and have lower pass rates are considered more difficult. At its core, difficulty lies in the size of the possible action space rather than in the quality of the data itself. More recently, LIMO \cite{ref12}, s1 \cite{ref37}, and related works adapt the Less-is-More hypothesis to reasoning, showing that a small set of high-quality and precise reasoning demonstrations suffices to obtain strong results. Some works \cite{ref17}, however, point to a relatively restrained ceiling when scaling the dataset size under this paradigm.

There is a gap between model performance and the knowledge a model actually contains \cite{ref32, ref34, ref39}, and an active direction aims to close this gap without further training or data scaling. These methods instead perform additional reasoning at test time (inference-time scaling), using extra compute to produce the final answer. \citet{ref38} find that scaling test-time computation can be more effective than scaling the size or training of the model under a similar compute budget. Depth-wise scaling methods such as chain-of-thought \cite{ref43} and reflection \cite{ref42, ref44} carefully extend the reasoning trajectory and can recover from early errors, making the process more stable. Width-wise methods, such as self-consistency \cite{ref46} and best-of-$N$ sampling, invoke the model's inner diversity to probe its ability ceiling, and further aggregation of multiple outputs can yield additional gains \cite{ref52, ref4}. Knowing when and how to scale remains an open question \cite{ref45, ref21}.

\subsection{The Use of Large Language Models (LLMs)}
We used Overleaf\cite{ref65} to improve writing and Cursor \cite{ref66} to help code.

\subsection{Token Categories}
\label{token-categories}
We categorize tokens into five groups based on their semantic roles (we category them with the meta-cognition method, by simply asking the LLMs to categorize them, and refine the classifications):

\noindent\textbf{Logical:} logical, reasoning, therefore, consequently, implies, follows that, deduce, infer, conclusion, premise, argument, valid, sound, if...then, necessary, sufficient, contradiction, fallacy

\noindent\textbf{Reflection:} reflect, think about, reconsider, rethink, looking back, in retrospect, upon reflection, wait, hold on, on second thought, reconsidering

\noindent\textbf{Mathematical:} calculate, compute, equation, formula, theorem, proof, mathematical, numerical, algebraic, geometric, arithmetic, derivative, integral, function, variable, constant, solve, solution, answer is, equals, sum, product

\noindent\textbf{Uncertainty:} maybe, perhaps, possibly, might, could be, uncertain, not sure, unclear, ambiguous, doubt, question, probably, likely, seems

\noindent\textbf{Emphasis:} definitely, certainly, absolutely, clearly, obviously, undoubtedly, without doubt, surely, indeed, in fact, especially, particularly, notably, importantly, key


\subsection{SOTA Comparison Table}

\begin{table*}[t]
\centering
\vspace{1em}
\setlength{\tabcolsep}{1.8mm}{
\begin{tabular}{l|c|c|c|c|c|c|c}
\hline
\textbf{Model}& \textbf{data} & \textbf{train} & \textbf{Strategy} & \textbf{A24} & \textbf{A25} & \textbf{MATH} & \textbf{GPQA} \\
\hline
LIMO-32B                     & 1k & SFT & - & 56.7 & 49.3 & 86.6 & 66.7 \\
S1-32B                       & 1k & SFT & - & 36.0 & 25.3 & 84.8 & 59.6 \\
S1.1-32B                     & 1k & SFT & - & 64.7 & 49.3 & 89.0 & 63.6 \\
\hline
OpenThinker-32B              & 114k & SFT & - & 68.0 & 49.3 & 90.6 & 63.5 \\
OpenThinker2-32B              & 1000K & SFT & - & 76.7 & 58.7 & 90.8 & 64.1 \\
Skywork-OR1-32B-Preview      & 124K & RL & - & 77.1 & 68.2 & 97.5 & - \\
MiroMind-M1-RL-32B           & 62K & RL & - & 77.5 & 65.6 & 96.4 & - \\
QwQ-32B                      & - & RL & - & 79.5 & 69.5 & 98.0 & - \\
Qwen3-32B                    & - & - & - & 81.4 & 72.9 & 97.2 & - \\
Qwen3-235B-A22B              & - & - & - & 85.7 & 81.5 & 98.0 & - \\
OpenAI-o1                    & - & - & - & 74.3 & 79.2 & 96.4 & - \\
OpenAI-o3-mini(medium)       & - & - & - & 79.6 & 74.8 & 98.0 & - \\
Grok-3-Beta(Think)           & - & - & - & 83.9 & 77.3 & - & - \\
DeepSeek-R1                  & - & - & - & 79.8 & 70.0 & -  & - \\
DeepSeek-Llama-70B& 800K & SFT & - & 70.0 & 56.3 & 94.5 & - \\
\hline
DeepSeek-Qwen-32B & 800K & SFT & - & 70.8 & 52.1 & - & - \\
DeepSeek-Qwen-32B & \multirow{2}{*}{800K} & \multirow{2}{*}{SFT} & \multirow{2}{*}{-} & \multirow{2}{*}{56.7} & \multirow{2}{*}{40.0} & \multirow{2}{*}{90.0} & \multirow{2}{*}{50.0} \\
(Greedy Decoding)   &  &  &  &  &  & \\
\hline
\multirow{7}{*}{Qwen2.5-32B-Instruct}&  &  & S-D & 76.7 & 70   & 95.0 & 69.7 \\
&  &  & D-S & 70   & 73.3 & 94.2 & 70.2 \\
&  &  & Merge & 70   & 76.7 & 95.0 & 68.7 \\
& 1K & SFT & Majority Vote & \multirow{2}{*}{73.3} & \multirow{2}{*}{80.0} & \multirow{2}{*}{94.2} & \multirow{2}{*}{70.2} \\
&  &  & (S-D D-S R-R) & & & & \\
&  &  & Majority Vote & \multirow{2}{*}{76.7} & \multirow{2}{*}{76.7} & \multirow{2}{*}{95.2} & \multirow{2}{*}{70.8} \\
&  &  & (S-D D-S R-S) & & & & \\
\hline
\multirow{7}{*}{DeepSeek-Qwen-32B}&  &  & D-S & 86.7&  \textbf{80.0}& 95.2 & 70.2\\
&  &  & S-S & \textbf{90.0} & 73.3& 94.4 & 67.8\\
&  &  & R-D & 83.3& 73.3& 96.2 & 66.2\\
& 1K & SFT & Majority Vote & \multirow{2}{*}{90.0} & \multirow{2}{*}{80.0} & \multirow{2}{*}{94.8} & \multirow{2}{*}{67.8} \\
&  &  & (D-S S-S S-S) & & & & \\
&  &  & Majority Vote & \multirow{2}{*}{90.0} & \multirow{2}{*}{80.0} & \multirow{2}{*}{96.4} & \multirow{2}{*}{68.2} \\
&  &  & (R-S D-R S-S) & & & & \\
\hline
\end{tabular}}
\caption{\textbf{Comparing the Performance of SOTA models} This table comparison showing pass@1 percentages across different models. We annotate the training dataset sizes, and their training methods to provide more details on how our method excels. The rows with a filled Strategy column are our models under specific train--test strategy combinations.}
\label{main-comparison}
\end{table*}

\subsection{More Experiments}
\textbf{Simply scaling the dataset size with the same quality on OpenThough dataset variants}

\begin{table*}[!htb]
\centering
\vspace{1em}
\setlength{\tabcolsep}{0.7mm}{
\begin{tabular}{c|c|cccccc|cccccc}
\hline
 &  & \multicolumn{6}{c|}{\textbf{OT-15k Dataset}} & \multicolumn{6}{c}{\textbf{OT-1k Dataset}} \\
\cline{3-14}
\textbf{Train} & \textbf{Test} & \textbf{A24} & \textbf{A25} & \textbf{MATH} & \textbf{GPQA} & \textbf{Avg} & \textbf{Avg2} & \textbf{A24} & \textbf{A25} & \textbf{MATH} & \textbf{GPQA} & \textbf{Avg} & \textbf{Avg2} \\
\hline
\multirow{4}{*}{\textbf{N}}
 & \textbf{N} & 46.67 & 50.00 & 89.80 & 52.02 & 59.62 & 48.34 & 63.67 & 50.00 & 94.80 & 66.67 & 68.79 & 56.84 \\
 & \textbf{R} & 40.00 & 36.67 & 79.20 & 53.54 & 52.35 & 38.34 & 60.00 & 53.33 & 91.20 & 66.67 & 67.80 & 56.67 \\
 & \textbf{S} & 50.00 & 30.00 & 79.80 & 43.56 & 50.84 & 40.00 & 60.00 & 56.67 & 90.00 & 64.85 & 67.88 & 58.34 \\
 & \textbf{D} & 53.33 & 50.00 & 79.20 & 45.45 & 56.99 & 51.67 & 60.00 & 60.00 & 89.40 & 61.62 & 67.76 & 60.00 \\
\hline
\multirow{4}{*}{\textbf{R}}
 & \textbf{N} & 45.67 & 23.33 & 88.40 & 55.56 & 53.24 & 34.50 & 40.00 & 26.67 & 92.80 & 68.18 & 56.91 & 33.34 \\
 & \textbf{R} & 50.00 & 43.33 & 92.00 & 44.44 & 57.44 & 46.67 & 60.00 & \textbf{70.00} & 94.80 & 69.70 & 73.63 & 65.00 \\
 & \textbf{S} & 63.33 & 53.33 & 92.40 & 45.05 & 63.53 & 58.33 & \textbf{73.33} & 53.33 & 94.00 & 67.33 & 72.00 & 63.33 \\
 & \textbf{D} & 60.00 & 66.67 & 91.20 & 47.98 & 66.46 & 63.34 & 66.67 & 56.67 & 94.40 & 69.70 & 71.86 & 61.67 \\
\hline
\multirow{4}{*}{\textbf{S}}
 & \textbf{N} & 52.33 & 30.00 & 88.80 & 56.57 & 56.93 & 41.17 & 43.33 & 33.33 & 91.60 & 68.69 & 59.24 & 38.33 \\
 & \textbf{R} & 63.33 & 56.67 & 89.00 & 44.44 & 63.36 & 60.00 & 73.33 & 60.00 & 94.40 & \textbf{71.72} & 74.86 & 66.67 \\
 & \textbf{S} & 53.33 & 56.67 & 87.80 & 46.53 & 61.08 & 55.00 & 66.67 & 66.67 & 94.60 & 69.31 & 74.31 & 66.67 \\
 & \textbf{D} & 46.67 & 50.00 & 89.60 & 35.35 & 55.41 & 48.34 & \textbf{76.67} & \textbf{70.00} & 95.00 & 66.67 & 77.09 & 73.34 \\
\hline
\multirow{4}{*}{\textbf{D}}
 & \textbf{N} & 46.33 & 33.33 & 88.40 & 51.52 & 54.90 & 39.83 & 33.33 & 36.67 & 91.80 & 66.16 & 56.99 & 35.00 \\
 & \textbf{R} & 50.00 & 43.33 & 91.40 & 45.45 & 57.55 & 46.67 & 63.33 & 66.67 & 94.60 & \textbf{72.22} & 74.21 & 65.00 \\
 & \textbf{S} & 66.67 & 46.67 & 90.20 & 45.54 & 62.27 & 56.67 & \textbf{70.00} & \textbf{73.33} & 94.20 & 67.82 & 76.34 & 71.67 \\
 & \textbf{D} & 66.67 & 60.00 & 90.80 & 43.43 & 65.23 & 63.34 & 66.67 & 60.00 & 94.20 & 67.68 & 72.14 & 63.34 \\
\hline
\end{tabular}}
\caption{Combined OT Dataset Results. We found that scaling the dataset size to 15k behaves much worse than only using 1K data. With on average 15\% performance down.}
\label{tab:ot_combined}
\end{table*}

Prior work~\citep{ref17} suggests that scaling dataset size can further improve performance. To test this, we scale OT from 1K to 15K examples drawn from the same distribution. Table~\ref{tab:ot_combined} shows that OT-15K \textbf{underperforms OT-1K by $\sim$15\%} on average: its best configuration (D-D: 66.7\%/60.0\% on AIME24/25) falls far short of OT-1K's best (S-D: 76.7\%/70.0\%), with no clear co-design pattern emerging---consistent with Finding~(b) that naively scaling quantity degrades performance.

\textbf{Merge \& Majority Vote results}\newline
\label{merge_majority}
Table~\ref{tab:merge} presents model merging results: merging Qwen2.5-32B-Instruct variants (S, D, R) via mergekit yields more stable performance (76.7\%/76.7\% on AIME24/25). Table~\ref{tab:majority} presents majority vote results across different strategy combinations. On Qwen2.5-32B-Instruct, majority vote achieves 80.0\%/76.7\% on AIME24/25; on DeepSeek-R1-Distill-Qwen-32B, majority vote reaches 90.0\%/80.0\%.
\begin{table}[!htb]
\centering
\caption{Merge Results}
\vspace{1em}
\label{tab:merge}
\setlength{\tabcolsep}{3.5mm}{
\begin{tabular}{l|l|c|c|c|c} 
\hline
\textbf{Train} & \textbf{Test} & \textbf{AIME24} & \textbf{AIME25} & \textbf{MATH} & \textbf{GPQA} \\
\hline
\multirow{4}{*}{\textbf{Merge}} 
    & \textbf{N} & 40.0 & 43.3 & 91.4 & 65.7 \\
    & \textbf{R} & 70.0 & \textbf{76.7} & 95.0 & 70.7 \\
    & \textbf{S} & 70.0 & 70.0 & 94.2 & \textbf{73.8} \\
    & \textbf{D} & \textbf{76.7} & 63.3 & 93.8 & 68.2 \\
\hline
\end{tabular}}
\end{table}
\begin{table}[!htb]
\centering
\caption{Majority vote results}
\vspace{1em}
\label{tab:majority}
\setlength{\tabcolsep}{3.5mm}{
\begin{tabular}{l|c|c|c} 
\hline
\textbf{Majority vote} & \textbf{AIME24} & \textbf{AIME25} & \textbf{MATH} \\
\hline
S-D D-S R-R & 0.733 & \textbf{0.800} & 0.942 \\
S-D D-S R-S & \textbf{0.767} & 0.767 & 0.952 \\
S-D D-S D-S & 0.733 & 0.733 & 0.956 \\
S-S D-D R-D & 0.667 & 0.600 & 0.952 \\
\hline
\end{tabular}}
\end{table}

\subsection{Reduced Benchmark-Gap Evidence}
\textbf{Observation.} All baseline methods in Table~\ref{main-comparison} exhibit substantial AIME24$>$AIME25 performance gaps: LIMO (7.4), S1.1 (15.4), OpenThinker2 (18.0), DeepSeek-R1 (9.8), QwQ (10.0). In contrast, our Input-Time Scaling configurations reduce this gap to 6.7 points or less while achieving higher absolute scores on both benchmarks.

\textbf{Analysis.} Both AIME 2024 and 2025 target the same AMC/AIME competition difficulty tier, ruling out genuine difficulty differences as the explanation. The gap appears systematically across all independent methods (none designed by us), making topic distribution effects implausible. The most parsimonious explanation is benchmark contamination: earlier problems are more likely to appear in pretraining corpora, creating inflated AIME24 scores. This pattern---earlier benchmarks scoring higher, with later benchmarks gradually catching up as models are retrained---is a well-documented phenomenon in the LLM evaluation literature.

\textbf{Our results.} After applying Input-Time Scaling, the gap shrinks dramatically (D-S: 6.7 on DeepSeek, $-3.3$ on Qwen2.5). Notably, our best-performing strategy combinations tend to be complementary (e.g., S-D, D-S)---out-of-distribution configurations that would not benefit from memorized benchmark patterns. This is consistent with our hypothesis that noise-augmented training forces models to learn general reasoning patterns rather than shortcuts tied to specific benchmark distributions.


\subsection{Prompts \& Domains \& Example Personas}
You can find the exact prompts, processes and persona examples in our supplementary files as  mentioned before. Below are some specific demonstrations to fit in the paper body.

\label{prompt}

The prompt for \textbf{Persona-Dissimilar} \newline
\colorbox{gray!20}{
  \parbox{\dimexpr\linewidth - 2\fboxsep\relax}{
    You should think of a persona based on the instructions provided. The output should only contain the persona without any irrelevant information.\newline instruction:[\textbf{INSTRUCTION}]\newline you should create a valid persona, instead of follow the instruction!! you should only generate the persona and don't do any calculation and reasoning!!! the persona must not contain any specific words from the instruction. And you should creatively name the person. The persona you thought of should not contain the ability to finish the instruction; instead, the persona should be considered far irrelative to solve the problem.
  }%
}
\newline
e.g.\newline
\colorbox{gray!20}{
  \parbox{\dimexpr\linewidth - 2\fboxsep\relax}{
    **Persona Name:** Jasper Frost  \newline\newline**Profession and Interests:** Jazz saxophonist with an unyielding fascination for urban rooftop gardening. Spends weekends restoring vintage vinyl records and crafting miniature terrariums inside lightbulbs. Secretly believes houseplants communicate through subtle leaf vibrations. Dislikes spreadsheets, refuses to own a calculator, and once tried to compost a broken metronome.
  }%
}
\newline\newline\newline
The prompt for \textbf{Persona-Similar} \newline
\colorbox{gray!20}{
  \parbox{\dimexpr\linewidth - 2\fboxsep\relax}{
    You should think of a persona based on the instructions provided. The output should only contain the persona without any irrelevant information.\newline instruction:[\textbf{INSTRUCTION}]\newline you should create a valid persona, instead of follow the instruction!! you should only generate the persona and don't do any calculation and reasoning!!! the persona must not contain any specific words from the instruction. And you should creatively name the person.
  }%
}
\newline
e.g.\newline
\colorbox{gray!20}{
  \parbox{\dimexpr\linewidth - 2\fboxsep\relax}{
    **Persona Name:** Dr. Lila Voss  \newline**Expertise:** Probability Theorist and Logic Puzzle Enthusiast  \newline**Background:** Dr. Voss is a professor of discrete mathematics with a knack for simplifying complex statistical scenarios. She thrives on crafting elegant solutions using combinatorial logic and frequently incorporates collaborative problem-solving dynamics into her research. Her passion lies in uncovering hidden symmetries in seemingly random events and mentoring students to approach challenges with methodical curiosity.
  }%
}
\newline\newline\newline\newline
The prompt for \textbf{Persona-Random} \newline
\colorbox{gray!20}{
  \parbox{\dimexpr\linewidth - 2\fboxsep\relax}{
    You should think of a persona based on the domain provided. The output should only contain the persona without any irrelevant information.\newline domain:[\textbf{DOMAIN}]\newline And you should creatively name the person. You should not directly borrow words from the domain name; you should explore the domain and use the possible sub-domain instead
  }%
}
e.g.\newline
\colorbox{gray!20}{
  \parbox{\dimexpr\linewidth - 2\fboxsep\relax}{
    **Persona Name:** Azure Mirage  \newline**Domain:** Ephemeral Artistry  \newline **Traits:** A whimsical creator who crafts transient sculptures from fog, melting ice, and vanishing ink. Obsessed with impermanence, they reject preservation, believing beauty exists only in fleeting moments. Their most famous work, *Whispers of Frost*, dissolved before dawn, surviving solely in witnesses’ hazy memories. Known for wearing iridescent scarves that shift colors mid-conversation, leaving observers unsure if they glimpsed reality or illusion.
  }%
}
\newline\newline\newline\newline
The prompt for \textbf{extracting domains using meta-cognition}, similar to those in \cite{ref47} \newline
\colorbox{gray!20}{
  \parbox{\dimexpr\linewidth - 2\fboxsep\relax}{
    First, consider the following definitions:\newline- Knowledge: A set of information, facts, descriptions, and  theories acquired through education of experience.\newline- Topic: A subset of knowledge that focuses on a particular area or subject.\newline Now, I am creating a dataset of engaging human knowledge that demonstrate the capabilities and value of human beings across the knowledge we perceive. Please help me generate high-quality and well coverages topics (in snake case) for this dataset. You should provide the answers with the format answer-explanation.\newline At last, you need give me an analyze on the topics you provided, and explain why these topics together can be a whole representation of human knowledge.
  }%
}
The \textbf{domains} extracted \newline
\colorbox{gray!20}{
  \parbox{\dimexpr\linewidth - 2\fboxsep\relax}{
    1. **Natural Sciences**: Physics, chemistry, biology, earth sciences, and astronomy — foundational understanding of the physical universe and its laws. \newline
2. **Formal Sciences**: Mathematics, logic, statistics, and theoretical frameworks (e.g., computational theory) — tools for abstract reasoning and problem-solving. \newline
3. **Applied Sciences \& Technology**: Engineering, robotics, AI, nanotechnology, and sustainable tech — practical innovations that reshape society and industry. \newline
4. **Social Systems**: Governance, economics, law, education, and urbanization — structures that organize human collaboration and societal progress. \newline
5. **Philosophy \& Ethics**: Metaphysics, epistemology, moral philosophy, and political theory — frameworks for critical thinking and value-driven decision-making.\newline
6. **History \& Cultural Evolution**: Civilizations, conflicts, technological diffusion, and cultural exchange — lessons from humanity’s collective past. \newline
7. **Environmental Systems**: Ecology, climate science, sustainability, and resource management — interdependence of humans and planetary health. \newline
8. **Health \& Medicine**: Disease prevention, genetics, mental health, and healthcare systems — advancing physical and psychological well-being. \newline
9. **Language \& Communication**: Linguistics, semiotics, translation, and digital media — tools for expressing ideas and bridging cultural divides. \newline
10. **Creative Expression**: Visual arts, literature, music, culinary arts, and design — cultural identity and innovation through aesthetic forms. \newline
11. **Space \& Exploration**: Astrophysics, planetary science, and interstellar travel — expanding humanity’s physical and intellectual frontiers. \newline
12. **Psychology \& Behavior**: Cognition, emotion, social dynamics, and neurodiversity — understanding individual and collective human behavior. \newline
13. **Digital Revolution**: Cybersecurity, blockchain, quantum computing, and data science — adaptation to the transformative power of information. \newline
14. **Ritual\& Belief Systems**: Religion, spirituality, festivals, and rites of passage — symbolic practices that shape meaning and community. 
  }%
}

\subsection{Tables \& Charts}
\label{full-results}
\begin{figure*}[!htb]
    \centering
    \begin{subfigure}[b]{0.48\textwidth}
        \centering
        \includegraphics[width=\textwidth]{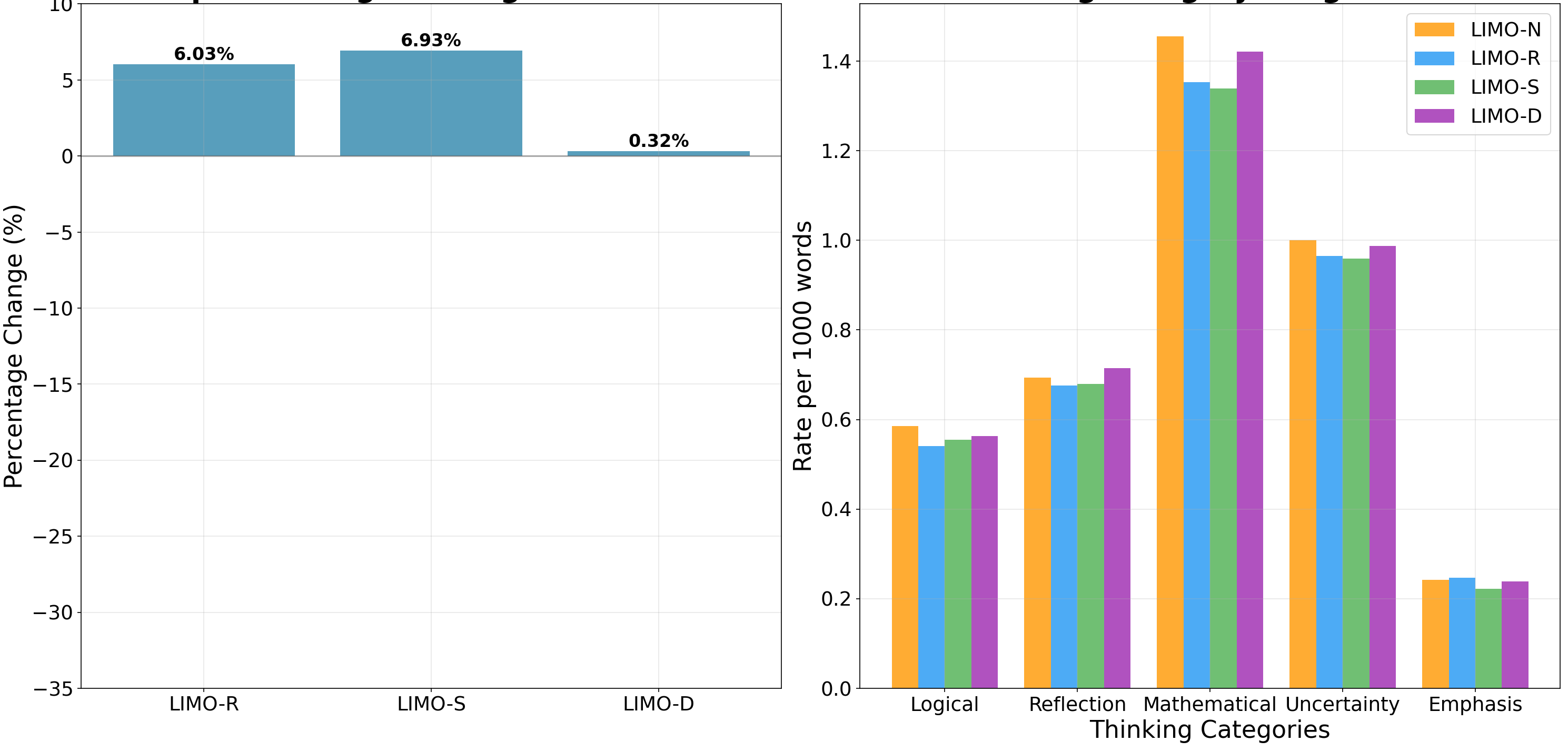}
        \caption{training, LIMO datasets.}
    \end{subfigure}
    \hfill
    \begin{subfigure}[b]{0.48\textwidth}
        \centering
        \includegraphics[width=\textwidth]{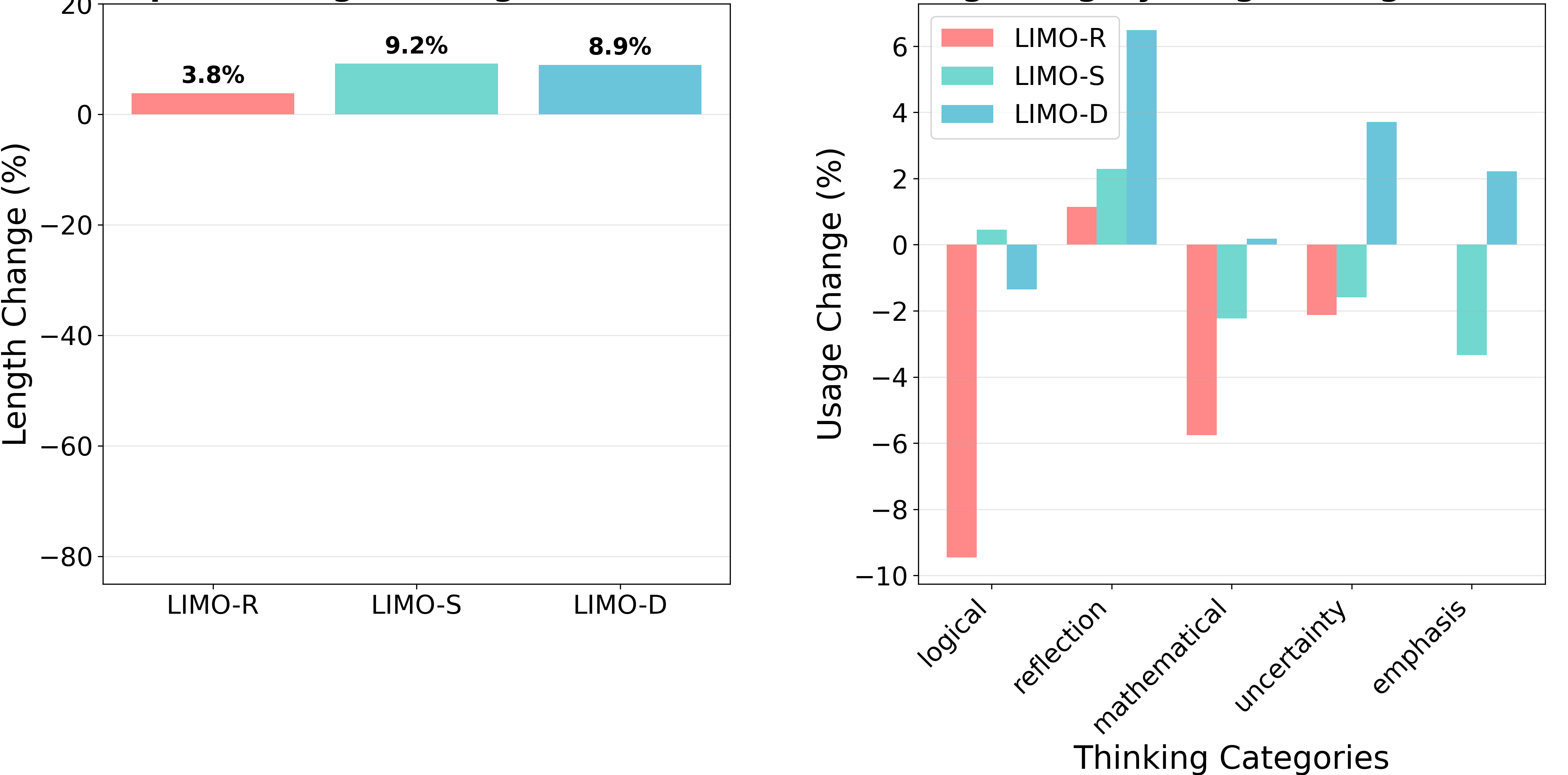}
        \caption{testing, LIMO datasets.}
    \end{subfigure}
    
    \vspace{0.3cm}
    
    \begin{subfigure}[b]{0.48\textwidth}
        \centering
        \includegraphics[width=\textwidth]{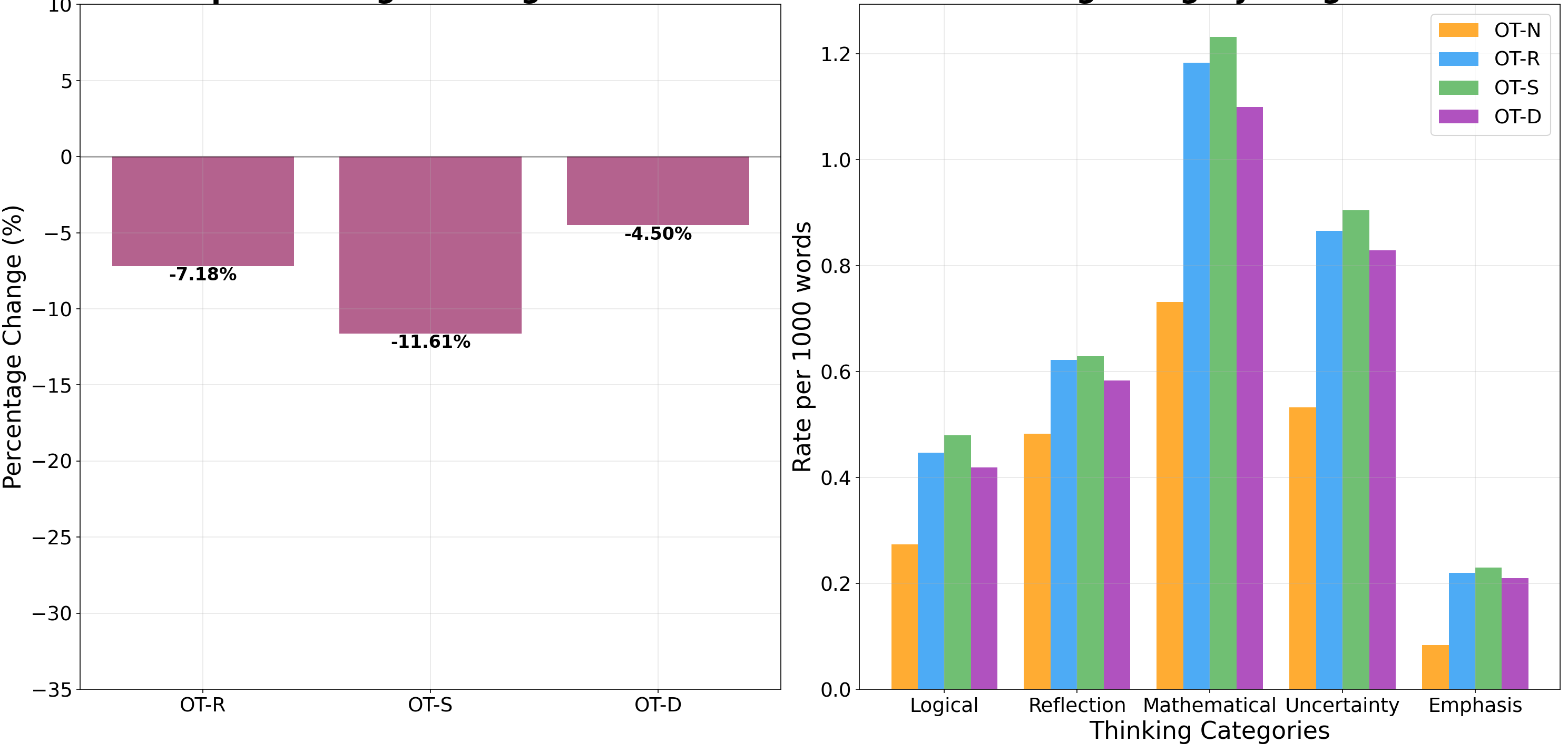}
        \caption{training, OT datasets.}
        \label{fig:ot_analysis}
    \end{subfigure}
    \hfill
    \begin{subfigure}[b]{0.48\textwidth}
        \centering
        \includegraphics[width=\textwidth]{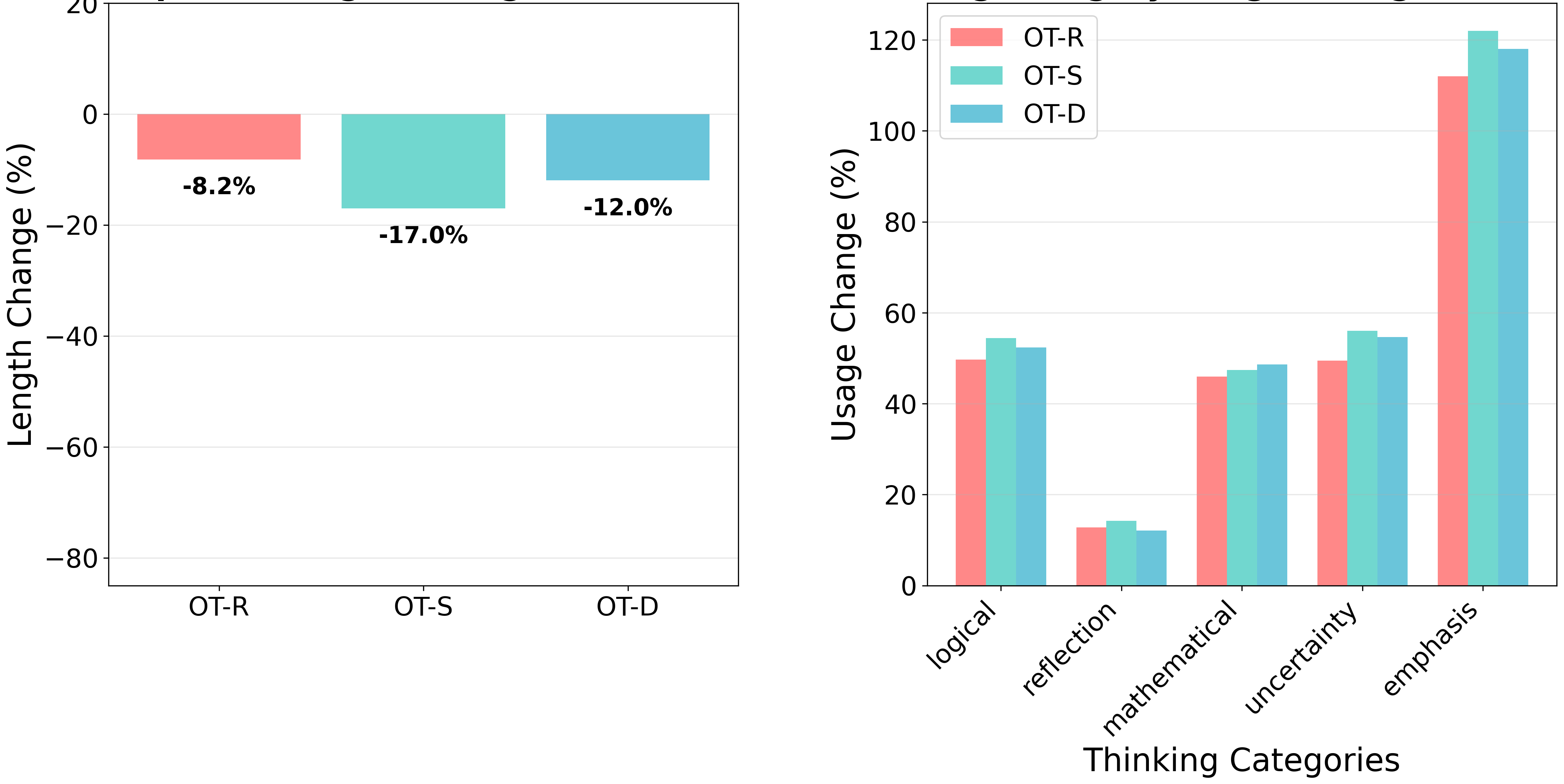}
        \caption{testing, OT datasets.}
        \label{fig:ot_test}
    \end{subfigure}
    
    \caption{Performance comparison between training and testing strategies of DeepSeek-R1-Distill-qwen-32B: (a,c) is the training visualization; (b,d) is the testing visualization.}
    \label{fig:deepseek-32b}
\end{figure*}

\begin{figure*}[!htbp]
    \centering
    \begin{subfigure}[b]{0.48\textwidth}
        \centering
        \includegraphics[width=\textwidth]{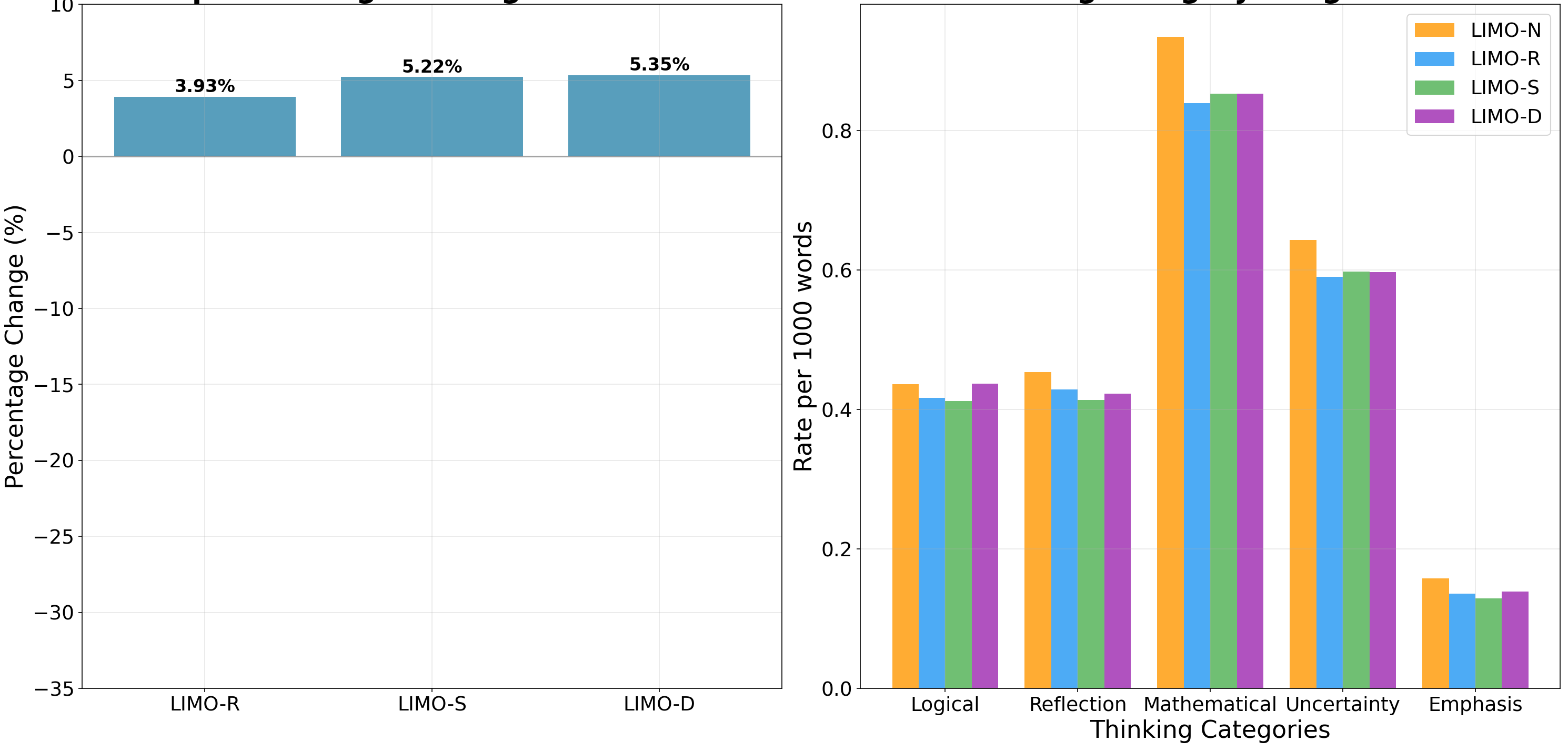}
        \caption{Qwen2.5-7B-Instruct, LIMO datasets.}
    \end{subfigure}
    \hfill
    \begin{subfigure}[b]{0.48\textwidth}
        \centering
        \includegraphics[width=\textwidth]{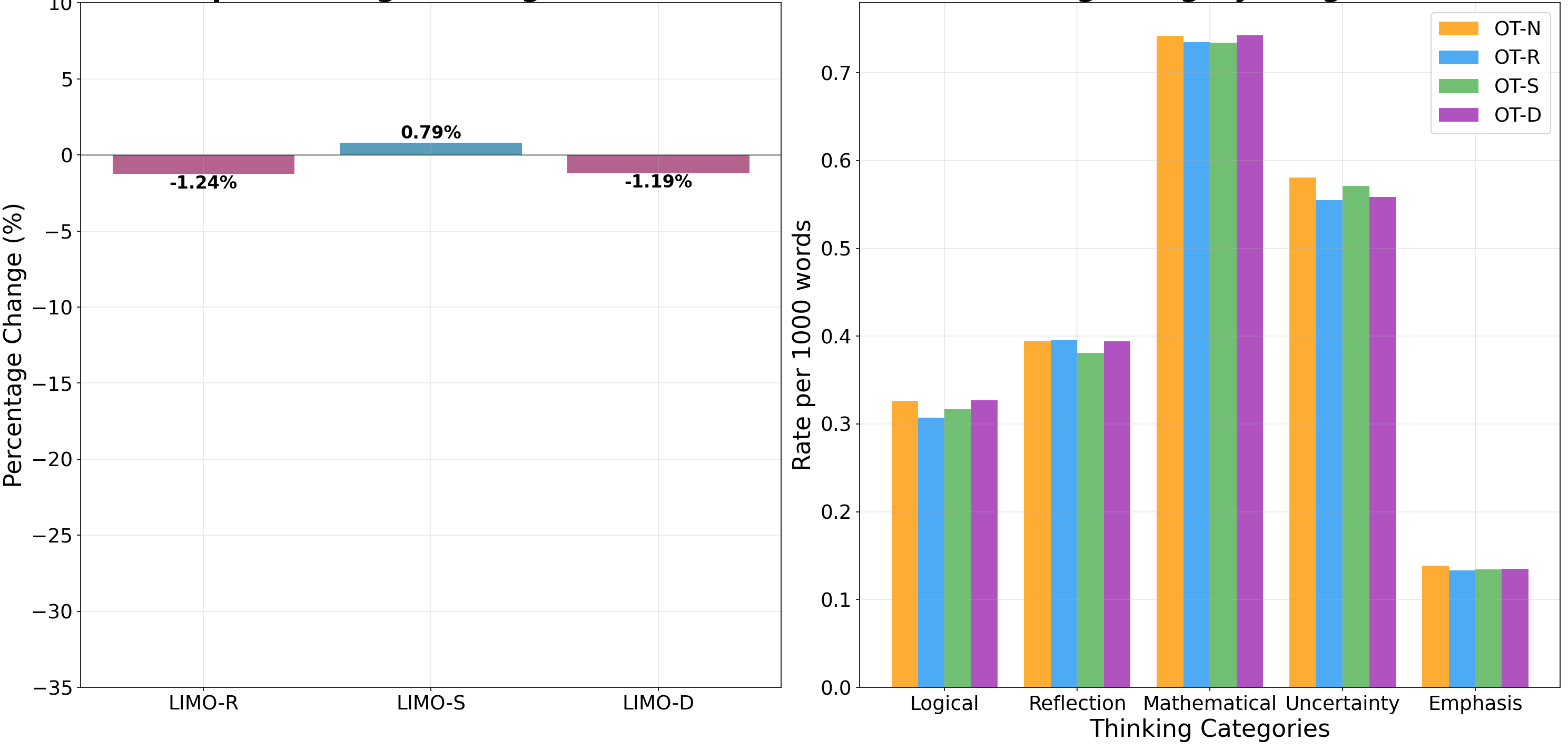}
        \caption{Qwen2.5-7B-Instruct, OT datasets.}
    \end{subfigure}
    
    \vspace{0.3cm}
    
    \begin{subfigure}[b]{0.48\textwidth}
        \centering
        \includegraphics[width=\textwidth]{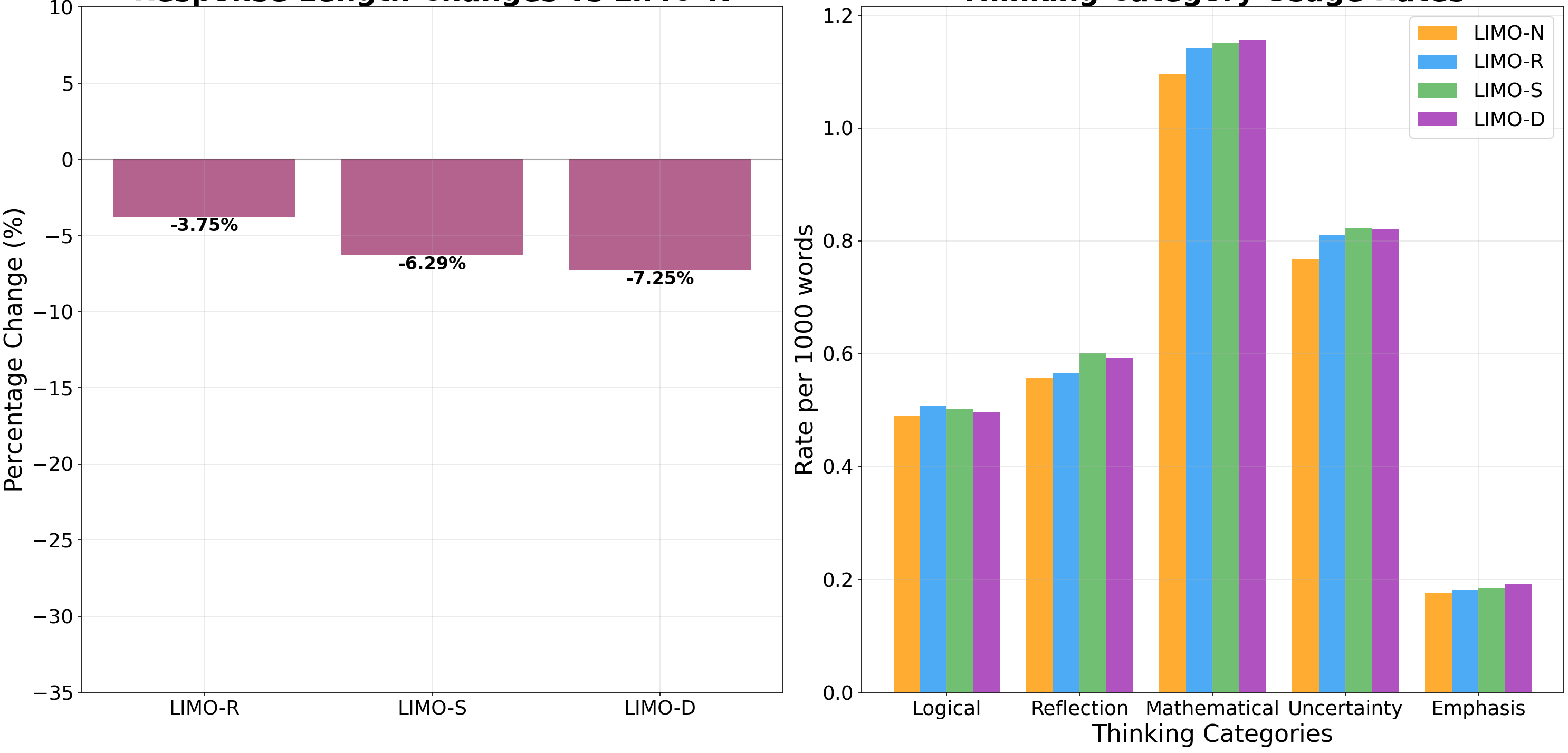}
        \caption{DeepSeek-R1-Distill-qwen-7B, LIMO datasets.}
    \end{subfigure}
    \hfill
    \begin{subfigure}[b]{0.48\textwidth}
        \centering
        \includegraphics[width=\textwidth]{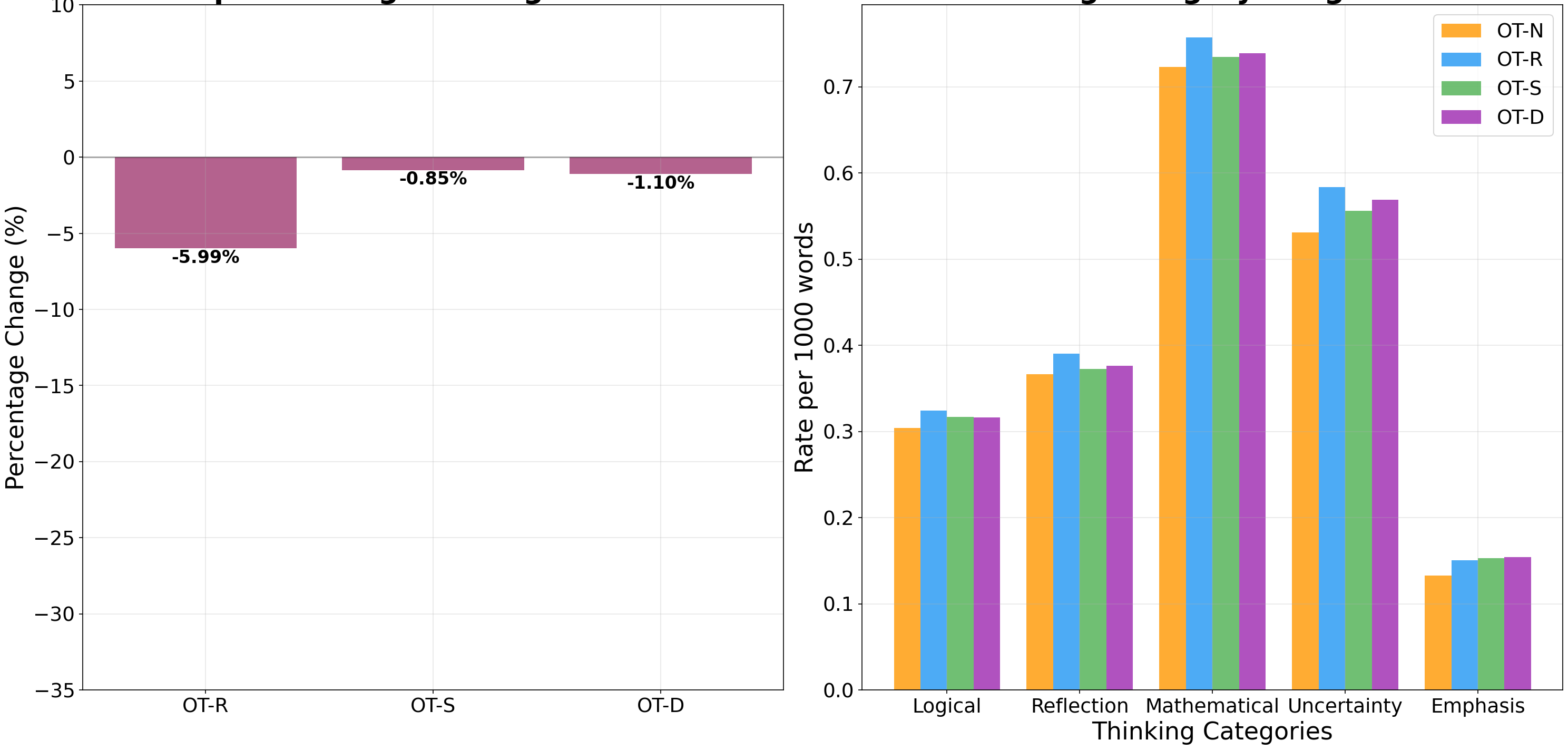}
        \caption{DeepSeek-R1-Distill-qwen-7B, OT datasets.}
    \end{subfigure}
    
    \caption{Performance comparison between training strategies}
    \label{fig:combined_comparison}
\end{figure*}

This section contains full 4$\times$4 training-testing strategy grids for all eight models, providing the complete evidence base for our findings. Each table follows the same format as Table~\ref{tab:qwen2.5-32b} in the main paper. Tables~\ref{tab:deepseek-r1-distill-qwen-1.5b}--\ref{tab:deepseek-r1-distill-qwen-32b} cover DeepSeek-R1-Distill-Qwen (1.5B, 7B, 32B); Tables~\ref{tab:qwen2.5-3b}--\ref{tab:qwen2.5-32b-2} cover Qwen2.5-Instruct (3B, 7B, 32B); Tables~\ref{tab:llama3-3b}--\ref{tab:llama3-8b} cover Llama (3B, 8B). Together, these tables support Finding~(a) (co-design consistency across all models), Finding~(b) (quality-capacity tradeoff), and Finding~(e) (emergent scaling at 32B). Figures~\ref{fig:deepseek-32b} and~\ref{fig:combined_comparison} visualize the token distribution patterns for 7B and 32B models, supporting Finding~(c) (intrinsic efficiency).

\begin{table*}[!htb]
\centering
\setlength{\tabcolsep}{0.7mm}{
\begin{tabular}{c|c|cccccc|cccccc}
\hline
 &  & \multicolumn{6}{c|}{\textbf{LIMO (High Quality)}} & \multicolumn{6}{c}{\textbf{OT (Low Quality)}} \\
\cline{3-14}
\textbf{Train} & \textbf{Test} & \textbf{A24} & \textbf{A25} & \textbf{MATH} & \textbf{GPQA} & \textbf{avg} & \textbf{avg2} & \textbf{A24} & \textbf{A25} & \textbf{MATH} & \textbf{GPQA} & \textbf{avg} & \textbf{avg2} \\
\hline
\multirow{4}{*}{\textbf{N}}
 & \textbf{N} & 7.50 & 6.67 & 69.40 & 5.05 & 22.15 & 7.08 & 10.00 & 11.67 & 42.80 & 1.01 & 16.37 & 10.84 \\
 & \textbf{R} & 11.67 & 11.67 & 67.00 & 3.03 & 23.34 & 11.67 & 14.17 & 14.17 & \textbf{53.60} & \textbf{2.53} & \textbf{21.12} & 14.17 \\
 & \textbf{S} & 10.83 & 5.83 & 72.00 & 1.49 & 22.54 & 8.33 & 17.50 & 11.67 & 48.00 & 1.49 & 19.66 & 14.59 \\
 & \textbf{D} & 9.17 & 12.50 & 72.00 & 3.54 & 24.30 & 10.84 & 11.67 & \textbf{15.83} & 50.00 & 2.02 & 19.88 & 13.75 \\
\hline
\multirow{4}{*}{\textbf{R}}
 & \textbf{N} & 13.33 & 12.50 & 70.80 & \textbf{7.07} & 25.93 & 12.91 & 12.50 & 13.33 & 41.60 & 0.51 & 16.98 & 12.91 \\
 & \textbf{R} & 10.83 & 8.33 & 70.80 & 5.05 & 23.75 & 9.58 & 10.83 & 10.00 & 45.00 & \textbf{2.53} & 17.09 & 10.41 \\
 & \textbf{S} & 14.17 & 9.17 & \textbf{74.80} & 4.95 & 25.77 & 11.67 & 14.17 & 12.50 & 40.60 & 0.00 & 16.82 & 13.34 \\
 & \textbf{D} & \textbf{16.67} & 10.00 & 74.20 & 5.05 & 26.48 & 13.33 & 15.00 & 12.50 & 42.00 & 1.01 & 17.63 & 13.75 \\
\hline
\multirow{4}{*}{\textbf{S}}
 & \textbf{N} & 8.33 & 10.00 & 71.00 & 3.54 & 23.22 & 9.16 & 15.00 & 14.17 & 42.40 & 1.01 & 18.14 & 14.59 \\
 & \textbf{R} & 12.50 & 10.00 & 73.40 & 6.57 & 25.62 & 11.25 & 14.17 & 15.00 & 43.60 & 0.51 & 18.32 & 14.59 \\
 & \textbf{S} & 8.33 & 10.83 & \textbf{74.80} & 2.97 & 24.23 & 9.58 & \textbf{21.67} & 10.83 & 38.20 & 0.50 & 17.80 & \textbf{16.25} \\
 & \textbf{D} & 15.00 & \textbf{18.33} & 73.40 & 5.05 & \textbf{27.95} & \textbf{16.66} & 15.00 & 13.33 & 47.40 & 0.51 & 19.06 & 14.16 \\
\hline
\multirow{4}{*}{\textbf{D}}
 & \textbf{N} & \textbf{16.67} & 15.00 & 71.20 & 5.56 & 27.11 & 15.83 & 15.00 & 10.83 & 43.00 & 0.51 & 17.34 & 12.91 \\
 & \textbf{R} & 14.17 & 7.50 & 73.00 & 6.06 & 25.18 & 10.84 & 17.50 & 9.17 & 42.20 & 1.01 & 17.47 & 13.34 \\
 & \textbf{S} & 10.00 & 15.00 & 73.00 & 5.45 & 25.86 & 12.50 & 8.33 & 9.17 & 42.60 & 0.99 & 15.27 & 8.75 \\
 & \textbf{D} & 14.17 & 10.83 & 73.80 & 6.06 & 26.21 & 12.50 & 15.83 & 12.50 & 44.40 & 1.52 & 18.56 & 14.16 \\
\hline
\end{tabular}}
\caption{\textbf{DeepSeek-R1-Distill-Qwen-1.5B:} "A24" is short for AIME24, and "A25" is short for AIME25. "Avg" is the average of four dataset results, and "Avg2" only counts for AIME24 and AIME25 results. We get the overall best performance with S-D under LIMO datasets. Avg2 is almost the same, and OT can even have a better performance, however, Avg is dominated by the MATH score, and LIMO perform much better than OT.}
\label{tab:deepseek-r1-distill-qwen-1.5b}
\end{table*}

\begin{table*}[!htb]
\centering
\setlength{\tabcolsep}{0.7mm}{
\begin{tabular}{c|c|cccccc|cccccc}
\hline
 &  & \multicolumn{6}{c|}{\textbf{LIMO (High Quality)}} & \multicolumn{6}{c}{\textbf{OT (Low Quality)}} \\
\cline{3-14}
\textbf{Train} & \textbf{Test} & \textbf{A24} & \textbf{A25} & \textbf{MATH} & \textbf{GPQA} & \textbf{avg} & \textbf{avg2} & \textbf{A24} & \textbf{A25} & \textbf{MATH} & \textbf{GPQA} & \textbf{avg} & \textbf{avg2} \\
\hline
\multirow{4}{*}{\textbf{N}}
 & \textbf{N} & 64.17 & 42.50 & 92.40 & 32.32 & \textbf{57.85} & 53.34 & \textbf{56.67} & 38.33 & 84.40 & 18.18 & 49.39 & \textbf{47.50} \\
 & \textbf{R} & 60.83 & 44.17 & 88.00 & 23.23 & 54.06 & 52.50 & 52.50 & 38.33 & 88.40 & 20.20 & 49.86 & 45.41 \\
 & \textbf{S} & 55.83 & 42.50 & 90.00 & 23.76 & 53.02 & 49.16 & 50.00 & 39.17 & 84.40 & 19.80 & 48.34 & 44.59 \\
 & \textbf{D} & 60.83 & 45.83 & 88.40 & 24.24 & 54.83 & 53.33 & 51.67 & \textbf{40.83} & 85.60 & 21.72 & 49.95 & 46.25 \\
\hline
\multirow{4}{*}{\textbf{R}}
 & \textbf{N} & 60.83 & 38.33 & 91.80 & \textbf{32.83} & 55.95 & 49.58 & 49.17 & 34.17 & 85.40 & \textbf{27.27} & 49.00 & 41.67 \\
 & \textbf{R} & 57.50 & 45.00 & \textbf{93.00} & 27.27 & 55.69 & 51.25 & 46.67 & 37.50 & 88.00 & 19.70 & 47.97 & 42.09 \\
 & \textbf{S} & 57.50 & \textbf{48.33} & 90.80 & 29.70 & 56.58 & 52.91 & 47.50 & 34.17 & 84.40 & 23.76 & 47.46 & 40.84 \\
 & \textbf{D} & \textbf{65.00} & 42.50 & 91.20 & 30.30 & 57.25 & \textbf{53.75} & 45.83 & 34.17 & 86.20 & 25.25 & 47.86 & 40.00 \\
\hline
\multirow{4}{*}{\textbf{S}}
 & \textbf{N} & 55.83 & 45.00 & 90.60 & 29.80 & 55.31 & 50.41 & 49.17 & 30.83 & 89.40 & 23.23 & 48.16 & 40.00 \\
 & \textbf{R} & 55.83 & 40.83 & 91.20 & 27.27 & 53.78 & 48.33 & 52.50 & 32.50 & 88.40 & 23.74 & 49.29 & 42.50 \\
 & \textbf{S} & 55.00 & 45.00 & 92.00 & 23.27 & 53.82 & 50.00 & 47.50 & 35.83 & \textbf{89.60} & 26.73 & 49.91 & 41.66 \\
 & \textbf{D} & 59.17 & 44.17 & 90.60 & 27.78 & 55.43 & 51.67 & 47.50 & 36.67 & 87.60 & 23.23 & 48.75 & 42.09 \\
\hline
\multirow{4}{*}{\textbf{D}}
 & \textbf{N} & 58.33 & 44.17 & 91.00 & 28.28 & 55.45 & 51.25 & 54.17 & 35.83 & 87.20 & 24.24 & \textbf{50.36} & 45.00 \\
 & \textbf{R} & 61.67 & 41.67 & \textbf{93.00} & 24.75 & 55.27 & 51.67 & 48.33 & 37.50 & 86.60 & 22.22 & 48.66 & 42.91 \\
 & \textbf{S} & 55.83 & 45.83 & 91.20 & 24.26 & 54.28 & 50.83 & 47.50 & 33.33 & 86.40 & 18.81 & 46.51 & 40.41 \\
 & \textbf{D} & 55.00 & 44.17 & 89.80 & 26.26 & 53.81 & 49.59 & 46.67 & 35.83 & 88.20 & 23.74 & 48.61 & 41.25 \\
\hline
\end{tabular}}
\caption{\textbf{DeepSeek-R1-Distill-Qwen-7B:} "A24" is short for AIME24, and "A25" is short for AIME25. "Avg" is the average of four dataset results, and "Avg2" only counts for AIME24 and AIME25 results. We get the overall best performance with S-D under LIMO datasets. The difference on percentage is smaller than that on 1.5B models.}
\label{tab:deepseek-r1-distill-qwen-7b}
\end{table*}

\begin{table*}[!htb]
\centering
\setlength{\tabcolsep}{0.7mm}{
\begin{tabular}{c|c|cccccc|cccccc}
\hline
 &  & \multicolumn{6}{c|}{\textbf{LIMO (High Quality)}} & \multicolumn{6}{c}{\textbf{OT (Low Quality)}} \\
\cline{3-14}
\textbf{Train} & \textbf{Test} & \textbf{A24} & \textbf{A25} & \textbf{MATH} & \textbf{GPQA} & \textbf{avg} & \textbf{avg2} & \textbf{A24} & \textbf{A25} & \textbf{MATH} & \textbf{GPQA} & \textbf{avg} & \textbf{avg2} \\
\hline
\multirow{4}{*}{\textbf{N}}
 & \textbf{N} & \textbf{86.67} & \textbf{73.33} & 94.80 & 66.16 & \textbf{80.24} & \textbf{80.00} & 56.67 & 40.00 & 90.00 & 50.00 & 59.17 & 48.34 \\
 & \textbf{R} & 66.67 & 60.00 & 89.20 & 65.15 & 70.25 & 63.34 & 63.33 & 46.67 & 81.40 & 56.57 & 61.99 & 55.00 \\
 & \textbf{S} & 63.33 & 66.67 & 91.20 & 67.33 & 72.13 & 65.00 & 63.33 & 40.00 & 83.20 & 57.92 & 61.11 & 51.66 \\
 & \textbf{D} & 76.67 & 60.00 & 85.40 & 66.67 & 72.19 & 68.34 & 43.33 & 36.67 & 76.00 & 56.06 & 53.02 & 40.00 \\
\hline
\multirow{4}{*}{\textbf{R}}
 & \textbf{N} & 76.67 & 70.00 & 94.40 & \textbf{71.72} & 78.20 & 73.34 & 70.00 & \textbf{80.00} & 95.60 & 70.20 & 78.95 & 75.00 \\
 & \textbf{R} & 66.67 & 63.33 & \textbf{95.40} & 68.69 & 73.52 & 65.00 & 76.67 & 63.33 & 94.80 & 67.68 & 75.62 & 70.00 \\
 & \textbf{S} & 76.67 & 53.33 & 94.80 & 66.83 & 72.91 & 65.00 & 80.00 & 76.67 & 95.80 & 67.82 & 80.07 & 78.34 \\
 & \textbf{D} & 63.33 & 63.33 & 95.20 & 63.64 & 71.38 & 63.33 & 83.33 & 73.33 & \textbf{96.20} & 66.16 & 79.75 & 78.33 \\
\hline
\multirow{4}{*}{\textbf{S}}
 & \textbf{N} & 76.67 & 66.67 & 95.20 & 69.70 & 77.06 & 71.67 & 80.00 & 70.00 & \textbf{96.20} & 69.70 & 78.97 & 75.00 \\
 & \textbf{R} & 83.33 & 60.00 & 94.20 & 68.69 & 76.55 & 71.66 & 83.33 & 73.33 & 95.00 & 70.20 & 80.47 & 78.33 \\
 & \textbf{S} & 76.67 & 63.33 & 94.40 & 70.79 & 76.30 & 70.00 & \textbf{86.67} & 76.67 & 93.80 & 70.79 & \textbf{81.98} & \textbf{81.67} \\
 & \textbf{D} & 80.00 & 60.00 & 94.60 & 69.19 & 75.95 & 70.00 & 83.33 & 66.67 & 94.40 & 68.18 & 78.14 & 75.00 \\
\hline
\multirow{4}{*}{\textbf{D}}
 & \textbf{N} & 73.33 & 63.33 & 95.00 & 70.20 & 75.47 & 68.33 & 76.67 & 73.33 & 95.80 & 68.69 & 78.62 & 75.00 \\
 & \textbf{R} & 73.33 & 60.00 & 95.00 & 69.70 & 74.51 & 66.66 & 76.67 & 66.67 & 92.80 & 68.18 & 76.08 & 71.67 \\
 & \textbf{S} & 80.00 & 56.67 & 93.80 & 66.83 & 74.33 & 68.34 & 80.00 & 73.33 & 93.60 & \textbf{71.78} & 79.68 & 76.66 \\
 & \textbf{D} & 76.67 & 70.00 & 94.00 & 70.20 & 77.72 & \textbf{73.34} & 80.00 & 60.00 & 92.40 & 68.18 & 75.14 & 70.00 \\
\hline
\end{tabular}}
\caption{\textbf{DeepSeek-R1-Distill-Qwen-32B:} "A24" is short for AIME24, and "A25" is short for AIME25. "Avg" is the average of four dataset results, and "Avg2" only counts for AIME24 and AIME25 results. We get the overall best performance with S-S under OT. Currently OT datasets show better results considering avg1 and on avg2. With the model becoming more capable, lower-quality datasets can actually gain a higher performance ceiling. While high-quality datasets are more restrained to Persona-N strategies.}
\label{tab:deepseek-r1-distill-qwen-32b}
\end{table*}

\begin{table*}[!htb]
\centering
\setlength{\tabcolsep}{0.7mm}{
\begin{tabular}{c|c|cccccc|cccccc}
\hline
 & & \multicolumn{6}{c|}{\textbf{LIMO (High Quality)}} & \multicolumn{6}{c}{\textbf{OT (Low Quality)}} \\
\cline{3-14}
\textbf{Train} & \textbf{Test} & \textbf{A24} & \textbf{A25} & \textbf{MATH} & \textbf{GPQA} & \textbf{avg} & \textbf{avg2} & \textbf{A24} & \textbf{A25} & \textbf{MATH} & \textbf{GPQA} & \textbf{avg} & \textbf{avg2} \\
\hline
\multirow{4}{*}{\textbf{N}}
 & \textbf{N} & 2.50 & 0.83 & 33.00 & 8.59 & 11.23 & 1.67 & 0.83 & 0.83 & 1.60 & 3.54 & 1.70 & 0.83 \\
 & \textbf{R} & 3.33 & 0.83 & 28.80 & 9.60 & 10.64 & 2.08 & 1.67 & 0.83 & 20.60 & 4.04 & 6.78 & 1.25 \\
 & \textbf{S} & 2.50 & 0.00 & 26.80 & \textbf{9.90} & 9.80 & 1.25 & 2.50 & 0.00 & 9.80 & 6.93 & 4.81 & 1.25 \\
 & \textbf{D} & 0.83 & 1.67 & 23.00 & 6.06 & 7.89 & 1.25 & 1.67 & 3.33 & 11.20 & 2.53 & 4.68 & \textbf{2.50} \\
\hline
\multirow{4}{*}{\textbf{R}}
 & \textbf{N} & 0.83 & 1.67 & \textbf{37.00} & 9.09 & \textbf{12.15} & 1.25 & 1.67 & 0.00 & \textbf{39.60} & \textbf{9.09} & \textbf{12.59} & 0.83 \\
 & \textbf{R} & 1.67 & 0.00 & 32.20 & 5.05 & 9.73 & 0.83 & 0.83 & 0.83 & 1.80 & 0.00 & 0.86 & 0.83 \\
 & \textbf{S} & 1.67 & 0.83 & 28.40 & 6.93 & 9.46 & 1.25 & \textbf{3.33} & 0.83 & 1.60 & 0.00 & 1.44 & 2.08 \\
 & \textbf{D} & 0.00 & 0.83 & 28.60 & 5.56 & 8.75 & 0.41 & 0.00 & 0.83 & 0.60 & 0.00 & 0.36 & 0.41 \\
\hline
\multirow{4}{*}{\textbf{S}}
 & \textbf{N} & 1.67 & 0.00 & 31.60 & 8.59 & 10.46 & 0.83 & 0.00 & 0.00 & 5.40 & 7.07 & 3.12 & 0.00 \\
 & \textbf{R} & 3.33 & 0.00 & 32.20 & 8.08 & 10.90 & 1.67 & 1.67 & 0.00 & 0.80 & 0.51 & 0.74 & 0.83 \\
 & \textbf{S} & 5.00 & 0.83 & 31.20 & 5.94 & 10.74 & \textbf{2.92} & 1.67 & 0.00 & 1.20 & 1.49 & 1.09 & 0.83 \\
 & \textbf{D} & 0.83 & 0.83 & 25.80 & 5.56 & 8.26 & 0.83 & 0.83 & \textbf{4.17} & 1.60 & 1.01 & 1.90 & \textbf{2.50} \\
\hline
\multirow{4}{*}{\textbf{D}}
 & \textbf{N} & 2.50 & 2.50 & 34.00 & 6.06 & 11.27 & 2.50 & 1.67 & 0.00 & 35.40 & 8.08 & 11.29 & 0.83 \\
 & \textbf{R} & 0.83 & 0.00 & 28.80 & 3.54 & 8.29 & 0.41 & 0.00 & 0.00 & 1.40 & 0.00 & 0.35 & 0.00 \\
 & \textbf{S} & 3.33 & 0.83 & 28.80 & 2.97 & 8.98 & 2.08 & 0.83 & 0.00 & 1.20 & 0.99 & 0.76 & 0.41 \\
 & \textbf{D} & 0.83 & 1.67 & 29.80 & 5.05 & 9.34 & 1.25 & 0.00 & 1.67 & 0.60 & 0.51 & 0.69 & 0.83 \\
\hline
\end{tabular}}
\caption{\textbf{Llama-3.2-3B:} "A24" is short for AIME24, and "A25" is short for AIME25. "Avg" is the average of four dataset results, and "Avg2" only counts for AIME24 and AIME25 results. It shows a clear dominance of MATH learning abilities on LIMO than OT datasets.}
\label{tab:llama3-3b}
\end{table*}

\begin{table*}[!htb]
\centering
\setlength{\tabcolsep}{0.7mm}{
\begin{tabular}{c|c|cccccc|cccccc}
\hline
 &  & \multicolumn{6}{c|}{\textbf{LIMO (High Quality)}} & \multicolumn{6}{c}{\textbf{OT (Low Quality)}} \\
\cline{3-14}
\textbf{Train} & \textbf{Test} & \textbf{A24} & \textbf{A25} & \textbf{MATH} & \textbf{GPQA} & \textbf{avg} & \textbf{avg2} & \textbf{A24} & \textbf{A25} & \textbf{MATH} & \textbf{GPQA} & \textbf{avg} & \textbf{avg2} \\
\hline
\multirow{4}{*}{\textbf{N}}
 & \textbf{N} & 2.50 & 4.17 & 58.80 & 37.88 & 25.84 & 3.33 & 4.17 & 2.50 & 44.20 & 25.25 & 20.07 & 3.33 \\
 & \textbf{R} & 9.17 & 1.67 & 55.40 & 40.40 & 26.66 & 5.42 & 5.83 & 5.00 & 43.00 & 23.74 & 19.18 & 5.42 \\
 & \textbf{S} & 7.50 & 3.33 & 56.60 & 41.09 & 27.13 & 5.42 & \textbf{10.00} & 7.50 & 43.40 & 27.72 & 21.59 & \textbf{8.75} \\
 & \textbf{D} & 8.33 & 7.50 & 50.80 & 38.38 & 26.25 & 7.92 & 5.00 & 5.00 & 40.00 & 21.72 & 19.60 & 5.00 \\
\hline
\multirow{4}{*}{\textbf{R}}
 & \textbf{N} & 3.33 & 1.67 & 59.40 & 37.88 & 25.57 & 2.50 & 4.17 & 1.67 & 48.20 & 26.26 & 20.07 & 2.92 \\
 & \textbf{R} & 6.67 & 5.00 & 59.20 & 40.40 & 27.82 & 5.83 & 9.17 & 6.67 & 47.00 & 23.74 & \textbf{21.64} & 7.92 \\
 & \textbf{S} & 6.67 & 4.17 & 59.20 & \textbf{45.54} & \textbf{28.89} & 5.42 & 6.67 & 5.00 & 46.20 & 28.22 & 21.52 & 5.83 \\
 & \textbf{D} & 5.00 & 5.83 & 56.60 & 39.39 & 26.70 & 5.42 & 5.83 & 5.83 & 41.40 & 22.22 & 18.82 & 5.83 \\
\hline
\multirow{4}{*}{\textbf{S}}
 & \textbf{N} & 10.83 & 7.50 & 71.80 & 7.58 & 24.43 & 9.16 & 6.67 & 4.17 & 49.20 & 25.76 & 21.45 & 5.42 \\
 & \textbf{R} & 11.67 & \textbf{10.83} & 71.00 & 5.05 & 24.64 & 11.25 & 7.50 & 3.33 & 49.60 & 22.73 & 20.79 & 5.42 \\
 & \textbf{S} & 13.33 & 10.00 & \textbf{74.20} & 3.96 & 25.37 & 11.66 & 6.67 & 4.17 & 47.80 & 27.72 & 21.59 & 5.42 \\
 & \textbf{D} & \textbf{15.00} & 9.17 & 72.00 & 6.06 & 25.56 & \textbf{12.09} & 6.67 & 7.50 & \textbf{52.80} & 19.19 & 21.54 & 7.08 \\
\hline
\multirow{4}{*}{\textbf{D}}
 & \textbf{N} & 9.17 & 4.17 & 59.20 & 40.91 & 28.36 & 6.67 & 7.50 & 3.33 & 44.20 & 25.25 & 20.07 & 5.42 \\
 & \textbf{R} & 5.83 & 2.50 & 60.00 & 40.40 & 27.18 & 4.17 & 5.00 & 5.00 & 43.00 & 23.74 & 19.18 & 5.00 \\
 & \textbf{S} & 8.33 & 4.17 & 59.40 & 39.60 & 27.88 & 6.25 & 6.67 & 5.00 & 43.40 & \textbf{31.19} & 21.57 & 5.83 \\
 & \textbf{D} & 5.00 & 5.83 & 59.00 & 44.44 & 28.57 & 5.42 & 7.50 & \textbf{9.17} & 40.00 & 21.72 & 19.60 & 8.34 \\
\hline
\end{tabular}}
\caption{\textbf{Llama-3.1-8B:} "A24" is short for AIME24, and "A25" is short for AIME25. "Avg" is the average of four dataset results, and "Avg2" only counts for AIME24 and AIME25 results. The difference on scores between LIMO and OT datasets is smaller than 3B models. With the models becoming more capable, the advantage of high-quality datasets is shrinking.}
\label{tab:llama3-8b}
\end{table*}

\begin{table*}[!htb]
\centering
\setlength{\tabcolsep}{0.7mm}{
\begin{tabular}{c|c|cccccc|cccccc}
\hline
 &  & \multicolumn{6}{c|}{\textbf{LIMO (High Quality)}} & \multicolumn{6}{c}{\textbf{OT (Low Quality)}} \\
\cline{3-14}
\textbf{Train} & \textbf{Test} & \textbf{A24} & \textbf{A25} & \textbf{MATH} & \textbf{GPQA} & \textbf{avg} & \textbf{avg2} & \textbf{A24} & \textbf{A25} & \textbf{MATH} & \textbf{GPQA} & \textbf{avg} & \textbf{avg2} \\
\hline
\multirow{4}{*}{\textbf{N}}
 & \textbf{N} & 5.83 & 2.50 & 44.60 & 8.59 & 15.38 & 4.17 & 0.83 & 2.50 & 1.00 & 1.52 & 1.46 & 1.67 \\
 & \textbf{R} & 2.50 & 6.67 & 42.40 & 5.56 & 14.28 & 4.58 & 4.17 & 2.50 & 6.40 & 0.51 & 3.40 & 3.33 \\
 & \textbf{S} & 5.83 & 5.00 & 42.40 & 5.94 & 14.79 & 5.42 & 2.50 & 2.50 & 3.80 & 1.49 & 2.57 & 2.50 \\
 & \textbf{D} & 5.83 & 3.33 & 37.60 & 6.57 & 13.33 & 4.58 & 4.17 & 5.00 & 3.40 & 0.51 & 3.27 & 4.58 \\
\hline
\multirow{4}{*}{\textbf{R}}
 & \textbf{N} & 5.83 & 5.83 & \textbf{49.40} & 6.06 & 16.78 & 5.83 & 5.00 & 2.50 & 63.20 & 18.18 & 22.22 & 3.75 \\
 & \textbf{R} & 4.17 & 5.83 & \textbf{49.40} & \textbf{9.60} & \textbf{17.25} & 5.00 & 4.17 & 7.50 & 2.40 & 1.52 & 3.90 & 5.83 \\
 & \textbf{S} & 2.50 & 5.83 & 47.00 & 9.41 & 16.18 & 4.17 & 5.00 & 5.83 & 3.20 & 1.98 & 4.00 & 5.42 \\
 & \textbf{D} & \textbf{7.50} & 6.67 & 44.40 & 6.57 & 16.29 & \textbf{7.08} & 5.00 & 5.83 & 2.80 & 1.01 & 3.66 & 5.42 \\
\hline
\multirow{4}{*}{\textbf{S}}
 & \textbf{N} & 3.33 & 3.33 & 46.40 & 9.09 & 15.54 & 3.33 & 5.83 & 5.83 & \textbf{65.20} & \textbf{24.24} & \textbf{25.28} & 5.83 \\
 & \textbf{R} & 4.17 & 2.50 & 45.80 & 8.59 & 15.27 & 3.33 & 4.17 & 3.33 & 3.20 & 0.00 & 2.68 & 3.75 \\
 & \textbf{S} & 5.00 & 6.67 & 43.40 & 7.43 & 15.62 & 5.83 & \textbf{6.67} & 2.50 & 2.00 & 0.00 & 2.79 & 4.58 \\
 & \textbf{D} & \textbf{7.50} & 2.50 & 44.00 & 5.56 & 14.89 & 5.00 & 4.17 & \textbf{10.83} & 2.60 & 0.51 & 4.53 & \textbf{7.50} \\
\hline
\multirow{4}{*}{\textbf{D}}
 & \textbf{N} & 3.33 & \textbf{7.50} & 47.80 & 7.58 & 16.55 & 5.42 & 4.17 & 6.67 & 62.00 & 19.70 & 23.13 & 5.42 \\
 & \textbf{R} & 5.00 & 6.67 & 47.00 & 4.04 & 15.68 & 5.83 & 5.00 & 2.50 & 3.60 & 0.51 & 2.90 & 3.75 \\
 & \textbf{S} & 6.67 & 5.00 & 48.60 & 4.46 & 16.18 & 5.83 & \textbf{6.67} & 1.67 & 6.60 & 1.49 & 4.11 & 4.17 \\
 & \textbf{D} & 5.00 & 6.67 & 47.20 & 8.08 & 16.74 & 5.83 & 2.50 & 5.00 & 5.60 & 0.51 & 3.40 & 3.75 \\
\hline
\end{tabular}}
\caption{\textbf{qwen2.5-3B:} "A24" is short for AIME24, and "A25" is short for AIME25. "Avg" is the average of four dataset results, and "Avg2" only counts for AIME24 and AIME25 results. LIMO datasets show a clear advantage on MATH scores, which dominates the avg.}
\label{tab:qwen2.5-3b}
\end{table*}

\begin{table*}[!htb]
\centering
\setlength{\tabcolsep}{0.7mm}{
\begin{tabular}{c|c|cccccc|cccccc}
\hline
 &  & \multicolumn{6}{c|}{\textbf{LIMO (High Quality)}} & \multicolumn{6}{c}{\textbf{OT (Low Quality)}} \\
\cline{3-14}
\textbf{Train} & \textbf{Test} & \textbf{A24} & \textbf{A25} & \textbf{MATH} & \textbf{GPQA} & \textbf{avg} & \textbf{avg2} & \textbf{A24} & \textbf{A25} & \textbf{MATH} & \textbf{GPQA} & \textbf{avg} & \textbf{avg2} \\
\hline
\multirow{4}{*}{\textbf{N}}
 & \textbf{N} & 18.33 & 15.83 & 77.80 & 35.35 & 36.82 & 17.08 & 22.50 & 21.67 & 37.00 & 5.05 & 21.55 & 22.09 \\
 & \textbf{R} & 18.33 & 14.17 & 74.20 & 27.78 & 33.62 & 16.25 & 17.50 & 21.67 & 36.80 & \textbf{8.59} & 21.14 & 19.59 \\
 & \textbf{S} & 20.00 & 14.17 & 77.40 & \textbf{40.59} & 38.04 & 17.09 & 24.17 & 17.50 & 35.00 & 5.94 & 20.65 & 20.84 \\
 & \textbf{D} & \textbf{22.50} & \textbf{24.17} & 74.20 & 37.37 & \textbf{39.56} & \textbf{23.34} & 21.67 & 21.67 & 27.00 & 6.06 & 19.10 & 21.67 \\
\hline
\multirow{4}{*}{\textbf{R}}
 & \textbf{N} & 18.33 & 16.67 & 77.40 & 32.83 & 36.31 & 17.50 & 24.17 & 17.50 & 43.00 & 2.53 & 21.80 & 20.84 \\
 & \textbf{R} & 20.00 & 19.17 & 77.20 & 33.33 & 37.42 & 19.59 & 20.83 & 20.83 & 44.40 & 6.06 & 23.03 & 20.83 \\
 & \textbf{S} & 20.83 & 17.50 & 77.20 & 29.70 & 36.31 & 19.16 & 19.17 & 21.67 & 44.00 & 3.47 & 22.08 & 20.42 \\
 & \textbf{D} & 15.00 & 21.67 & 76.00 & 31.82 & 36.12 & 18.34 & 22.50 & \textbf{25.00} & 45.60 & 5.05 & 24.54 & 23.75 \\
\hline
\multirow{4}{*}{\textbf{S}}
 & \textbf{N} & 20.83 & 18.33 & 77.80 & 36.87 & 38.46 & 19.58 & 25.83 & 20.00 & 48.20 & 7.58 & 25.40 & 22.91 \\
 & \textbf{R} & 18.33 & 17.50 & 76.40 & 30.30 & 35.63 & 17.91 & 25.00 & 19.17 & 45.40 & 6.06 & 23.91 & 22.09 \\
 & \textbf{S} & 18.33 & 18.33 & 77.60 & 30.20 & 36.12 & 18.33 & 24.17 & 17.50 & 43.20 & 3.96 & 22.21 & 20.84 \\
 & \textbf{D} & 15.83 & 18.33 & \textbf{79.00} & 31.82 & 36.24 & 17.08 & \textbf{26.67} & 22.50 & \textbf{48.60} & 6.57 & \textbf{26.09} & \textbf{24.59} \\
\hline
\multirow{4}{*}{\textbf{D}}
 & \textbf{N} & 20.00 & 17.50 & 77.00 & 37.37 & 37.97 & 18.75 & 21.67 & 19.17 & 44.60 & 3.54 & 22.25 & 20.42 \\
 & \textbf{R} & 14.17 & 17.50 & 77.80 & 31.31 & 35.20 & 15.84 & 23.33 & 16.67 & 39.80 & 2.53 & 20.58 & 20.00 \\
 & \textbf{S} & 21.67 & 16.67 & 78.00 & 32.67 & 37.25 & 19.17 & 24.17 & 23.33 & 44.40 & 3.96 & 23.96 & 23.75 \\
 & \textbf{D} & 16.67 & 18.33 & 78.80 & 26.77 & 35.14 & 17.50 & 20.83 & 23.33 & 42.00 & 3.54 & 22.43 & 22.08 \\
\hline
\end{tabular}}
\caption{\textbf{qwen2.5-7B:} "A24" is short for AIME24, and "A25" is short for AIME25. "Avg" is the average of four dataset results, and "Avg2" only counts for AIME24 and AIME25 results. LIMO datasets show a clear advantage on MATH scores, but the difference is becoming smaller. And OT shows a clear advantage on hard questions of AIME24 and AIME25.}
\label{tab:qwen2.5-7b}
\end{table*}

\begin{table*}[t]
\centering
\setlength{\tabcolsep}{0.7mm}{
\begin{tabular}{c|c|cccccc|cccccc}
\hline
 &  & \multicolumn{6}{c|}{\textbf{LIMO (High Quality)}} & \multicolumn{6}{c}{\textbf{OT (Low Quality)}} \\
\cline{3-14}
\textbf{Train} & \textbf{Test} & \textbf{A24} & \textbf{A25} & \textbf{MATH} & \textbf{GPQA} & \textbf{avg} & \textbf{avg2} & \textbf{A24} & \textbf{A25} & \textbf{MATH} & \textbf{GPQA} & \textbf{avg} & \textbf{avg2} \\
\hline
\multirow{4}{*}{\textbf{N}}
 & \textbf{N} & 61.00 & 40.00 & 92.00 & 53.54 & 61.64 & 50.50 & 63.67 & 50.00 & 94.80 & 66.67 & 68.79 & 56.84 \\
 & \textbf{R} & 43.33 & 40.00 & 90.20 & 51.01 & 56.14 & 41.67 & 60.00 & 53.33 & 91.20 & 66.67 & 67.80 & 56.67 \\
 & \textbf{S} & \textbf{63.33} & 40.00 & 89.00 & 50.00 & 60.58 & 51.67 & 60.00 & 56.67 & 90.00 & 64.85 & 67.88 & 58.34 \\
 & \textbf{D} & 40.00 & 36.67 & 87.20 & 51.52 & 53.85 & 38.34 & 60.00 & 60.00 & 89.40 & 61.62 & 67.76 & 60.00 \\
\hline
\multirow{4}{*}{\textbf{R}}
 & \textbf{N} & 59.33 & 36.67 & \textbf{93.80} & \textbf{68.69} & 64.62 & 48.00 & 40.00 & 26.67 & 92.80 & 68.18 & 56.91 & 33.34 \\
 & \textbf{R} & \textbf{63.33} & 50.00 & 93.20 & 65.66 & 68.05 & 56.67 & 60.00 & 70.00 & 94.80 & 69.70 & 73.63 & 65.00 \\
 & \textbf{S} & \textbf{63.33} & 53.33 & 93.40 & 66.34 & \textbf{69.10} & \textbf{58.33} & 73.33 & 53.33 & 94.00 & 67.33 & 72.00 & 63.33 \\
 & \textbf{D} & 53.33 & \textbf{56.67} & 92.40 & 63.13 & 66.38 & 55.00 & 66.67 & 56.67 & 94.40 & 69.70 & 71.86 & 61.67 \\
\hline
\multirow{4}{*}{\textbf{S}}
 & \textbf{N} & 60.33 & 46.67 & 92.20 & 59.09 & 64.57 & 53.50 & 43.33 & 33.33 & 91.60 & 68.69 & 59.24 & 38.33 \\
 & \textbf{R} & 43.33 & 43.33 & 90.20 & 58.08 & 58.74 & 43.33 & 73.33 & 60.00 & 94.40 & 71.72 & 74.86 & 66.67 \\
 & \textbf{S} & 60.00 & 50.00 & 92.00 & 56.44 & 64.61 & 55.00 & 66.67 & 66.67 & 94.60 & 69.31 & 74.31 & 66.67 \\
 & \textbf{D} & 53.33 & 50.00 & 89.40 & 59.09 & 62.96 & 51.67 & \textbf{76.67} & 70.00 & \textbf{95.00} & 66.67 & \textbf{77.09} & \textbf{73.34} \\
\hline
\multirow{4}{*}{\textbf{D}}
 & \textbf{N} & 55.67 & 43.33 & 92.20 & 58.59 & 62.45 & 49.50 & 33.33 & 36.67 & 91.80 & 66.16 & 56.99 & 35.00 \\
 & \textbf{R} & 60.00 & 50.00 & 89.00 & 53.03 & 63.01 & 55.00 & 63.33 & 66.67 & 94.60 & \textbf{72.22} & 74.21 & 65.00 \\
 & \textbf{S} & \textbf{63.33} & 46.67 & 90.60 & 57.43 & 64.51 & 55.00 & 70.00 & \textbf{73.33} & 94.20 & 67.82 & 76.34 & 71.67 \\
 & \textbf{D} & 53.33 & 33.33 & 88.00 & 52.02 & 56.67 & 43.33 & 66.67 & 60.00 & 94.20 & 67.68 & 72.14 & 63.34 \\
\hline
\end{tabular}}
\caption{\textbf{qwen2.5-32B:} "A24" is short for AIME24, and "A25" is short for AIME25. "Avg" is the average of four dataset results, and "Avg2" only counts for AIME24 and AIME25 results. Now the OT dataset gains both Avg and Avg2 higher than the LIMO dataset by a large margin. It improves more significantly on hard problems when the model capability is sufficient.}
\label{tab:qwen2.5-32b-2}
\end{table*}

\end{document}